\pdfoutput=1

\documentclass[11pt]{article}

\usepackage[final]{acl}
\usepackage{acl}
\usepackage{multirow}
\usepackage{amsmath}
\usepackage{fontenc}

\usepackage{times}
\usepackage{latexsym}
\usepackage{multirow}

\usepackage[most]{tcolorbox}
\usepackage{xcolor}
\usepackage{colortbl}
\definecolor{myorange}{HTML}{fa8072}
\definecolor{mygreen}{HTML}{9DC183}
\definecolor{lightgrey}{RGB}{158, 158, 158}
\definecolor{goldenrod}{rgb}{0,0,0.8}
\definecolor{deepred}{rgb}{0.6,0,0}
\definecolor{deepgreen}{rgb}{0,0.5,0}
\definecolor{pink}{RGB}{219, 48, 122}
\definecolor{forestgreen}{RGB}{34,139,34}
\definecolor{goldenrod}{RGB}{218,165,32}
\definecolor{sepia}{RGB}{112,66,20}
\usepackage{enumitem}
\usepackage{ulem}
\usepackage{multicol}
\usepackage{amssymb}
\usepackage{hyperref}
 \usepackage{mathtools}

\newcommand{\tick}{\textcolor{green!50!black}{\checkmark}}
\newcommand{\cross}{\textcolor{red}{\textsf{\small{x}}}}
\let\cite\citep

\usepackage[textsize=scriptsize]{todonotes}
\usepackage{xspace}

\makeatletter
\newcommand*\iftodonotes{\if@todonotes@disabled\expandafter\@secondoftwo\else\expandafter\@firstoftwo\fi}  
\makeatother

\definecolor{darkblack}{rgb}{0.0,0.0,0.5}
\definecolor{darkgreen}{rgb}{0.0, 0.42, 0.24}
\definecolor{lightgreen}{rgb}{0.52, 0.73, 0.4}
\definecolor{darkgray}{rgb}{0.4,0.4,0.4}
\definecolor{darkblue}{rgb}{0.0,0.0,0.5}
\definecolor{darkpurple}{rgb}{0.5,0.2,0.8}
\definecolor{lightpurple}{rgb}{0.8,0.5,1}

\usepackage{cleveref}
\crefname{figure}{Fig}{Figures}
\crefname{table}{Table}{Tables}
\crefname{appendix}{Appendix}{Apps.}
\crefname{section}{\S}{\S\S}
\crefformat{section}{\S#2#1#3} 
\crefname{equation}{Eq.}{Eqs.}
\crefname{algorithm}{Alg.}{Algs.}
\crefname{algocf}{Alg.}{Algs.}

\usepackage{wrapfig}
\usepackage{tabularx}
\usepackage{booktabs}
\usepackage{dsfont}
\usepackage{multirow}
\usepackage{array}

\usepackage{caption}
\usepackage{listings}
\usepackage{adjustbox}

\usepackage[table]{xcolor}
\usepackage{textcomp}
\newcommand{\mytexttilde}{\raisebox{0.5ex}{\texttildelow}}
\definecolor{TodoColor}{rgb}{1,0.7,0.6}
\definecolor{aquamarine}{rgb}{0.5, 1.0, 0.83}
\definecolor{some_blue}{rgb}{0.4, 0.4, 1.0}

\newcommand{\multirowcell}[1]{\begin{tabular}[c]{@{}c@{}}#1\end{tabular}}

\usepackage{tabularx}

\usepackage{listings}
\usepackage{todonotes}

\usepackage{subcaption}

\usepackage[T1]{fontenc}

\usepackage[utf8]{inputenc}

\usepackage{microtype}

\usepackage{inconsolata}

\usepackage{graphicx}

\usepackage{booktabs} 
\usepackage{diagbox}
\usepackage{CJK}
\usepackage{xspace} 
\usepackage{minted}
\usepackage{enumitem}
\usepackage{amssymb} 
\usepackage{cleveref} 
\usepackage{listings}
\usepackage{xcolor}
\usepackage{adjustbox}

\usepackage{xcolor}
\usepackage{listings}
\definecolor{mdtext}{rgb}{0,0,0}
\definecolor{mdback}{rgb}{0.95,0.95,0.95}
\definecolor{mdhead}{rgb}{0.2,0.2,0.7}

\definecolor{notebookbg}{RGB}{248,248,248} 

\newcommand{\xv}{\mathbf{x}}
\newcommand{\yv}{\mathbf{y}}
\newcommand{\rrv}{\mathbf{r}}

\newcommand{\uncerttext}[3]{%
  \def\sym{}%
  \ifdim #1 pt < 0.2pt
    \ifnum #3=0 \def\sym{\cross}%
    \else \def\sym{\tick}%
    \fi
    \tcbox[colback=green!20, boxrule=0pt, sharp corners, boxsep=0pt, left=1pt, right=1pt, top=0.5pt, bottom=0.5pt, on line]{#2~\raisebox{0.15em}{\scriptsize\sym}}%
  \else\ifdim #1 pt < 0.6pt
    #2%
  \else
    \ifnum #3=1 \def\sym{\cross}%
    \else \def\sym{\tick}%
    \fi
    \tcbox[colback=red!20, boxrule=0pt, sharp corners, boxsep=0pt, left=1pt, right=1pt, top=0.5pt, bottom=0.5pt, on line]{#2~\raisebox{0.15em}{\scriptsize\sym}}%
  \fi\fi
}

\tcbuselibrary{listingsutf8}

\lstdefinelanguage{markdown}{
  morecomment = [l][\color{mdhead}\bfseries]{\#},   
  morecomment = [l][\color{mdhead}]{>},             
  morecomment = [l][\color{mdhead}]{-},             
  morecomment = [l][\color{mdhead}]{*},             
  morestring  = [b][\color{mdcode}]{`},
  alsoletter  = {-},
  sensitive   = false
}

\lstdefinestyle{mdstyle}{
  language        = Markdown,
  basicstyle      = \ttfamily\small\color{mdtext},
  backgroundcolor = \color{mdback},
  frame           = single,
  rulecolor       = \color{black!15},
  breaklines      = true,
  showstringspaces= false,
  numbers         = none,
  morecomment     = [l][\color{mdquote}]{>},      
  morecomment     = [l][\color{mdhead}]{\#},      
  morecomment     = [l][\color{mdhead}]{-},       
}

\newcommand\myparagraph[1]{
\vskip 0.04in 
\noindent{\bf {#1}}}

\title{ReProbe: Efficient Test-Time Scaling of Multi-Step Reasoning by \\ Probing Internal States of Large Language Models}

\author{Jingwei Ni\textsuperscript{1 $\diamondsuit$} \, Ekaterina Fadeeva\textsuperscript{1 $\diamondsuit$} \, Tianyi Wu\textsuperscript{2 $\diamondsuit$}\\
\textbf{Mubashara Akhtar}\textsuperscript{1} \, \textbf{Jiaheng Zhang}\textsuperscript{2} \, \textbf{Elliott Ash}\textsuperscript{1} \, \textbf{Markus Leippold}\textsuperscript{4}\\ \textbf{Timothy Baldwin}\textsuperscript{3,5} \, \textbf{See-Kiong Ng}\textsuperscript{1} \, \textbf{Artem Shelmanov}\textsuperscript{3 $^{\blacklozenge}$} \, \textbf{Mrinmaya Sachan}\textsuperscript{1 $^{\blacklozenge}$} \\
\textsuperscript{1}ETH Z\"urich \, \textsuperscript{2}National University of Singapore \, \textsuperscript{3}MBZUAI \\ \textsuperscript{4}University of Z\"urich \, \textsuperscript{5}The University of Melbourne \\
$^{\diamondsuit}$Equal contribution \, $^{\blacklozenge}$Equal supervision}

\begin{document}

\maketitle

\begin{abstract}
LLMs can solve complex tasks by generating long, multi-step reasoning chains. Test-time scaling (TTS) can further improve performance by sampling multiple variants of intermediate reasoning steps, verifying their correctness, and selecting the best steps for continuation. However, existing verification approaches, such as Process Reward Models (PRMs), are computationally expensive and require large-scale human or model-generated annotations. We propose a lightweight alternative for step-level reasoning verification based on probing the internal states of LLMs.
We train a transformer-based probe that uses the internal states of a frozen LLM to estimate the credibility of its reasoning steps during generation. Annotation can be provided either by a larger LLM (e.g., DeepSeek-R1) or in a self-supervised manner by the original model itself. The probes are lightweight, containing fewer than 10M parameters. Across multiple domains, including mathematics, planning, and general knowledge question answering, our probes match or exceed the performance of PRMs that are up to 810× larger.
These results suggest that LLM internal states encode confidence in their reasoning processes and can serve as reliable signals for step verification, offering a promising path toward scalable, generalizable TTS and more introspective LLMs.\footnote{Code and data: \url{https://reprobe.github.io/}.}
\end{abstract}

\begin{figure}[t]
    \centering
    \includegraphics[width=1.0\columnwidth]{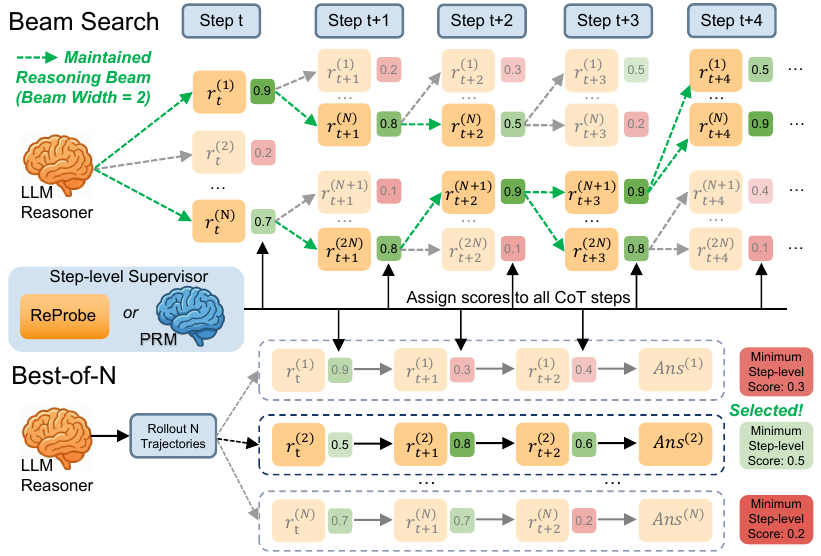}
    \caption{Illustration of the two prevalent test-time scaling methods: Best-of-N and Beam Search.}
    \label{fig:reprobe_tts}
\end{figure}

\section{Introduction}

Chain-of-thought (CoT) prompting has proven highly effective in eliciting the reasoning capabilities of Large Language Models (LLMs) to solve complex tasks \cite{wei2022chain}. Recent post-training approaches further enhance this capability through reinforcement learning, where models are rewarded for producing responses that demonstrate chain-of-thought reasoning prior to emitting the final answer, 
leading to the range of latest Large Reasoning Models (LRMs) \cite{deepseekai2025deepseekr1incentivizingreasoningcapability,yang2025qwen3,abdin2025phi}. 

Even the best LRMs generate incorrect reasoning trajectories and ultimately erroneous outputs. A single flawed reasoning step can derail the entire solution. Empirical evidence suggests, however, that 
allowing the model to attempt the task multiple times or exploring multiple candidate continuations at intermediate steps can increase the likelihood of reaching a correct answer -- a paradigm known as \textbf{Test-Time Scaling (TTS)} or test-time compute scaling \cite{yao2023tree,snell2024scaling}. Many TTS strategies rely on a scorer that evaluates the quality of individual reasoning steps and provides guidance to steer the reasoning process towards better solutions (see \Cref{fig:reprobe_tts}).

\begin{figure*}[t]
    \centering
    \includegraphics[width=1.0\linewidth]{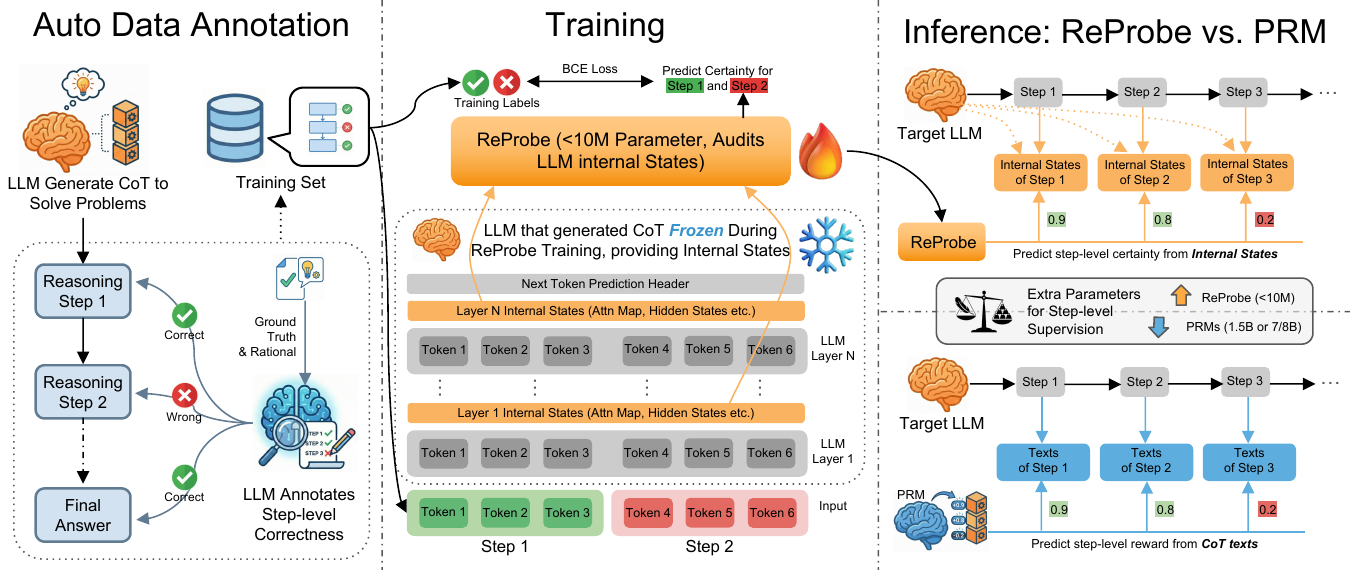}
    \caption{\textbf{Left}: Given a problem set, target LLM generates CoTs and an annotator LLM labels step-level correctness. \textbf{Middle}: Training ReProbe with target LLM providing internal states while frozen. \textbf{Right}: at inference, ReProbe and PRM monitors internal states and output text correspondingly. 
    }
    \label{fig:uq_vs_prm}
\end{figure*}

The contemporary approach to reasoning step verification is through process reward models (PRMs: \citealp{lightman2023letsverifystepstep,zhang2025lessonsdevelopingprocessreward}). Originally designed to guide RL-based post-training, PRMs were found to be particularly useful for TTS as well.
However, PRM-based verification faces key drawbacks. First, it requires expensive Monte-Carlo rollouts for training data annotation and a significant amount of computation for fine-tuning. Constructing training data is also complicated in tasks where step-level correctness cannot be inferred from the final result, such as mathematical proofs \cite{azerbayev2023proofnetautoformalizingformallyproving} and planning \cite{zheng2024naturalplanbenchmarkingllms}.
Second, deploying PRMs incurs substantial inference-time overhead, as they require running an additional LLM that significantly increases GPU memory usage and computational cost.
Third, PRMs are typically fine-tuned on narrow domains, such as mathematics, and therefore exhibit limited generalization. 

In another interesting line of work, the correctness of reasoning steps is assessed through unsupervised uncertainty quantification (UQ) 
\cite{gal2016uncertainty,malinin2020uncertainty,
vashurin2025cocoa,shelmanov-etal-2025-uncertainty}. Unlike PRMs, which represent external knowledge, UQ assumes that a model’s outputs and internal states provide information about the reliability of its generations \citep{fu2025deepthinkconfidence,yan2025mur,kang2025scalable}. 
While computationally lightweight, unsupervised UQ methods typically exhibit limited effectiveness in related tasks such as hallucination detection \cite{chuang-etal-2024-lookback,vazhentsev-etal-2025-token,vazhentsev-etal-2025-unconditional,shelmanov2025headpredictheadquestion}.



In this work, we investigate whether the cost-efficiency of UQ methods can be combined with the strong performance of PRMs. Prior work has shown that LLM internal states contain rich credibility signals that can be efficiently extracted with learned classifiers for hallucination detection \cite{azaria-mitchell-2023-internal,he-etal-2024-llm,shelmanov2025headpredictheadquestion}. Motivated by this observation, we introduce a lightweight training-based \textbf{Reasoning Probe (ReProbe)} that introspectively analyzes LLM internal states to assess the credibility of generated reasoning steps. In contrast to PRMs, which depend on external knowledge to evaluate reasoning traces, ReProbe relies only on model signals already available during generation: intermediate hidden states, attention weights, and logits -- thereby enabling substantially more efficient step-level reasoning verification while maintaining strong performance (see \Cref{fig:uq_vs_prm}).

Extensive in-domain (ID) and out-of-domain (OOD) evaluations highlight the advantages of our approach in three key aspects: \textbf{(1)~Strong generalizable performance}: ReProbes achieve competitive or superior performance compared to PRMs in step-level verification and test-time scaling, especially when dealing with OOD reasoning tasks; 
\textbf{(2)~Computational efficiency}: ReProbes outperform PRMs that are up to $150\times$ larger and remain competitive with PRMs up to $810\times$ larger. While typical PRMs contain 1.5--8B parameters, the probes trained in our work require fewer than 10M parameters; 
\textbf{(3) Training data efficiency}: unlike PRMs, which often rely on proprietary datasets \cite{skyworkopeno12024}, costly human annotations \cite{lightman2023letsverifystepstep}, or consensus-filtering involving both Monte Carlo rollouts and LLM judgments \cite{zhang2025lessonsdevelopingprocessreward}, ReProbes can be trained on data annotated automatically in a self-supervised and cost-efficient manner.

\section{Background}

\subsection{Test-Time Scaling}

LLMs solve complex tasks by generating 
CoT reasoning traces~\citep{guo2025deepseek}. Given an input question \(\xv\), models produce 
a sequence of reasoning steps \(\rrv = \{r_1, r_2, \dots, r_T\}\) followed by a final answer \(\yv=g(\xv,\rrv)\), and we assume that we can effectively extract individual steps and the final answer.  
Although linear CoT reasoning improves final task performance, locally incorrect steps \(r_t\) can propagate and cause incorrect answers.

\emph{Test-time scaling} \cite{yao2023tree,snell2024scaling} aims to further improve the performance of LLMs without retraining the entire model by allocating additional computation during inference or using more sophisticated inference strategies. By leveraging multiple candidate solutions and verification guidance, TTS facilitates the selection of better answers. Common approaches include best-of-$N$ (BoN) sampling and beam search \cite{xie2023self,yao2023tree,snell2024scalingllmtesttimecompute} (\cref{fig:reprobe_tts}). 

\myparagraph{Best-of-$N$ (BoN)} 
assumes that we sample not one, but $N$ reasoning trajectories $\{ \rrv^{(1)}, \rrv^{(2)}, \dots, \rrv^{(N)} \}$. Each trajectory $\rrv^{(j)}$ is evaluated using a scoring function $Q_{\text{BoN}}\!\left(\rrv^{(j)}\right)$, and the final answer is derived from the best-scoring trajectory:
\[
    \rrv^{*} = \arg\max_{1\leq j \leq N} Q_{\text{BoN}}\!\left(\rrv^{(j)}\right), \quad
    \yv=g(\xv,\rrv^{*}).
\]

\myparagraph{Beam Search (BS)} \cite{xie2023self,snell2024scalingllmtesttimecompute}, also known as the tree-of-thought with breadth-first search \cite{yao2023tree},
assumes that we keep $B \ge 1$ best reasoning trajectories so far, and for each of them, at a new step $t$, we sample $N$ variants of continuations: $\{ r_t^{(i,1)}, r_t^{(i,2)}, \dots, r_t^{(i,N)} \}, 1 \leq i \leq B$. Each of the continuations is assessed using a quality function $Q_{\text{bs}}\!\left( \rrv^{(i,k)}_{1:t}\right)$, and reasoning trajectories are expanded with the best continuations $\rrv^{(i)}_{1:t}=\rrv^{(i)}_{1:t-1} \circ r_t^{(i,*)}$.
After reaching the final step $T$, the answer is obtained from the best reasoning trajectory.
\[
S'_t
=
\left\{
\, \rrv^{(i)}_{1:t-1} \circ r_t^{(i,k)}
\;\middle|\;
\rrv^{(i)}_{1:t-1} \in S^*_{t-1},\;
1 \leq k \leq N
\right\}, 
\]
\begin{equation*}
\begin{aligned}
    S^{*}_t = 
    \arg\max_{S \subseteq S'_t,\; |S| = B}
    \sum_{\rrv^{(i,k)}_{1:t} \in S} Q_{\text{bs}}\!\left(\rrv_{1:t}^{(i,k)}\right), \nonumber \\
    \rrv^*=\arg\max_{\rrv^{(i)}_{1:T} \in S^*_T}Q_{\text{bs}}\left(\rrv^{(i)}_{1:T}\right), \; \yv=g(\xv,\rrv^{*}).
\end{aligned}
\end{equation*}

\subsection{Verifying Reasoning Steps} 
\label{subsec:scoring}


The most prominent approach for assessing reasoning steps in both BoN and beam search test-time scaling to date has been PRMs \cite{lightman2023letsverifystepstep,luo2024improvemathematicalreasoninglanguage,wang-etal-2024-math}. 
PRMs are critics designed to estimate the quality of each partial reasoning state 
$\rrv_{1:t}$ by assigning a reward 
$R_\text{PRM}\!\left(\rrv_{1:t}\right)$ that reflects the likelihood of the chain eventually 
leading to a correct solution. PRMs are typically implemented by a separate LLM 
trained to evaluate the plausibility and correctness of intermediate reasoning steps. The annotation for training PRMs comes from various sources, including crowdsourcing, self-consistency checking, and synthetic data generation pipelines. 



In BS, the reward from a PRM is used as a step-wise quality function. In BoN, the score for the complete chain $\rrv^{(j)}$ can be obtained via a temporal aggregation of process rewards, e.g., minimum step score \citep{zhang2025lessonsdevelopingprocessreward}:
\begin{equation*}
\begin{aligned}
Q_{\text{BoN}}\!\left(\rrv^{(j)}\right) 
\;=\; \min_{1 \leq t \leq T^{(j)}}  \nonumber
R_{\text{PRM}}\!\left(\rrv^{(j)}_{1:t}\right), \\ \nonumber
Q_{\text{bs}}\!\left(\rrv^{(i,k)}_{1:t}\right)=
R_{\text{PRM}}\left(\rrv^{(i,k)}_{1:t}\right).  \nonumber 
\end{aligned}
\end{equation*}



PRMs come with several limitations: (1) they are relatively large models, typically containing 1.5B-8B parameters, which leads to additional computational and memory overhead during inference; (2)~PRMs are often domain-specific, being trained for tasks such as mathematical reasoning, and exhibit limited generalization to unseen domains.

Recent work also suggests using confidence scores for the verification of reasoning steps or entire reasoning trajectories as a drop-in replacement for PRMs \cite{fu2025deepthinkconfidence}: the lower the uncertainty, the more trustworthy is the step. This is motivated by their efficiency and generalization. 

\section{Verifying Reasoning Steps via Probing Internal States of LLMs} \label{sec:method}

\myparagraph{ReProbe model.}
We construct a reasoning probe as a \emph{plug-and-play} module on top of a frozen LLM that outputs a probability of how likely the current reasoning step is correct. While PRMs rely solely on generated tokens, our probe introspectively leverages richer per-token features extracted from the base LLM’s internal states, which allows it to be both efficient and effective (see \Cref{fig:uq_vs_prm}).


The feature extractor $F(\xv, \rrv_{1:t-1}, r_{t})$ can leverage various internal signals at each token position. Specifically, we investigate two sets of features: (1)~\textbf{Attn+Logit}: attention weights over the 5 preceding tokens from all layers and logits of the top-K candidate generations; and (2) \textbf{Hidden states} from all layers. We select these feature sets because they have proven effective for factual hallucination detection \citep{shelmanov2025headpredictheadquestion,azaria-mitchell-2023-internal}. However, their role in reasoning supervision has been unexplored. 


The architecture of ReProbe is as follows:
\begin{enumerate}
    \item \textbf{Linear feature projection layer} that adjusts the dimension of token-level features:
    \[
        \mathbf{h}_0 = \mathrm{Linear}_{\mathrm{proj}}(F(\xv, \rrv_{1:t-1}, r_{t})).
    \]
    \item \textbf{Transformer layers:} a stack of $L$ layers that capture contextual dependencies:
    \[
        \mathbf{h}_l = \mathrm{Transformer}_l(\mathbf{h}_{l-1}), \quad l = 1, \dots, L.
    \]
    \item \textbf{Aggregation} of token-level features $h_L^{\tau}$ into a step-level vector:
    \[
        h_{r} = \frac{1}{|\rrv_t|}\sum_{\tau \in \rrv_t}h_L^{\tau}.
    \]
    \item \textbf{Two-layer classification NN} that outputs a final logit $\mathbf{z}$:
    \[
        \mathbf{z} = \mathrm{Linear}_2\!\Big(
            \sigma\big(\mathrm{Dropout}(\mathrm{Linear}_1(h_r))\big)
        \Big),
    \]
    where $\sigma(\cdot)$ is a GeLU activation function.
\end{enumerate}






\myparagraph{Reasoning step extraction} is done in two ways. LLMs in non-thinking mode are instructed to generate each CoT step on a separate, self-contained line (see \cref{appendix:llm_prompt}). LLMs reliably follow this format and produce well-formatted CoT steps, so they can be easily extracted during training and inference via simple pattern matching. LLMs operating in native thinking mode cannot be explicitly instructed to produce reasoning trajectories with clearly separated steps. To mitigate this problem, we treat each sentence as a reasoning step. 


\myparagraph{Constructing training data.}
To construct the training data for ReProbe, we use 10.8K problems (prompts) 
from the training subset of PRM800K. This dataset is an established resource for training and evaluating LLM reasoning capabilities. As most PRMs are already trained on LLM generations derived from it, it enables a fair comparison \citep{wang-etal-2024-math,zhang2025lessonsdevelopingprocessreward}. 
We generate 3 reasoning trajectories per problem across 10.8K mathematical problems, yielding \mytexttilde32K samples. We use nucleus sampling (top-$k{=}50$, top-$p{=}0.95$, temperature${=}1.0$). For annotation and training efficiency, we restrict the length of the generation during training. This restriction is removed during evaluation to ensure that our results are free of length bias.

We annotate the correctness of each reasoning step using an LLM as a judge. The judge is provided with the input question, the target LLM's CoT steps and final answer, as well as the ground-truth answer, and is prompted to assess the correctness of each reasoning step. The annotation prompts are provided in ~\cref{appendix:anno_prompt}. We consider two approaches to step annotation:
(1) an external verification setting, in which a larger LLM evaluates the reasoning steps -- following \citet{zheng2025processbenchidentifyingprocesserrors}, we use DeepSeek-R1; and (2) a self-supervised setting, in which the same LLM annotates its own generated CoT steps.
Other details for reproducing the training data construction can be found in \cref{appendix:training_data_anno}.

\myparagraph{Training ReProbe.} We train ReProbe with a standard binary cross-entropy objective and compensate for the severe class imbalance using class weighting. Only the probe parameters are updated, while the underlying LLM remains entirely frozen. This makes training highly efficient in practice: the total annotation cost is only \$200, and the required compute amounts to just 4 GH200 GPU-hours (see \cref{appendix:computation_for_training}). To further improve scalability, we implement hidden-state extraction and ReProbe training in a vLLM-based pipeline, which reduces end-to-end overhead in practice and is included in our released code.

\begin{table*}[ht]
\centering
\centering

\resizebox{\linewidth}{!}{
\begin{tabular}{lc|ccc|ccc|cc|ccc}
\toprule
\multirow{2}{*}{\textbf{Method}} & \multirow{2}{*}{\textbf{\# Sample}} & \multicolumn{3}{c|}{\textbf{Math (ID)}} & \multicolumn{3}{c|}{\textbf{Planning (OOD)}} & \multicolumn{2}{c|}{\textbf{QA (OOD)}} & \multicolumn{3}{c}{\textbf{Average}} \\
 &  & \textbf{MATH} & \textbf{GSM8k} & \textbf{ProofNet} & \textbf{Trips} & \textbf{Meetings} & \textbf{Calendar} & \textbf{StrQA} & \textbf{SciQA} & \textbf{ID} & \textbf{OOD} & \textbf{Overall}\\
\midrule
\rowcolor{gray!20}
 \multicolumn{13}{c}{\textit{Unsupervised Uncertainty Quantification (UQ)}} \\
\midrule
Random & - & .173 & .061 & .153 & .524 & .588 & .486 & .116 & .125 & .129 & .368 & .278 \\
MaxProb & - & .221 & .106 & .194 & .578 & .655 & .483 & .114 & .259 & .174 & .418 & .326 \\
MaxEntropy & - & .212 & .122 & .185 & .545 & .618 & .443 & .119 & .227 & .173 & \underline{.390} & .309 \\
Perplexity & - & .205 & .099 & .176 & .519 & .572 & .418 & .110 & .228 & .160 & .369 & .291 \\
Self-Certainty & - & .213 & .101 & .155 & .516 & .643 & .482 & .120 & .243 & .156 & .401 & .309 \\
CCP & - & .250 & .090 & .168 & .584 & .645 & .452 & .119 & .235 & .169 & .407 & .318 \\
P(True) & - & .164 & .059 & .172 & .535 & .608 & .490 & .126 & .263 & .132 & .404 & .302 \\
Semantic Entropy & - & .257 & .116 & .173 & .565 & .610 & .492 & .111 & .265 & .182 & .409 & .324 \\
Lexical Similarity & - & .250 & .119 & .170 & .569 & .603 & .490 & .120 & .254 & .180 & .407 & .322 \\
Degree Matrix & - & .227 & .089 & .147 & .534 & .597 & .484 & .107 & .258 & .154 & .396 & .305 \\
\midrule
\rowcolor{gray!20}
 \multicolumn{13}{c}{\textit{PRMs \textbf{150$\times$} Larger than ReProbes}} \\
\midrule
Skywork-PRM-1.5B & Unk & .283 & .412 & .147 & .433 & .532 & .502 & .254 & .408 & .281 & .426 & .371 \\
H4-Qwen2.5-PRM-1.5B-0.2 & 369K & .259 & .171 & .159 & .597 & .633 & .495 & .213 & .228 & .196 & .433 & .344 \\
\midrule
\rowcolor{gray!20}
 \multicolumn{13}{c}{\textit{PRMs \textbf{750$\times$} to \textbf{810$\times$} Larger than ReProbes}} \\
\midrule
Math-Shepherd-PRM-7B & 440K & .380 & .405 & .147 & .662 & .660 & .657 & .284 & .415 & .311 & .536 & .451 \\
RLHFlow-PRM-Deepseek-8B & 253K & .289 & .540 & .136 & .583 & .579 & .504 & .390 & \underline{.518} & .322 & .515 & .442 \\
RLHFlow-PRM-Mistral-8B & 273K & .233 & .537 & .118 & .523 & .555 & .499 & .349 & .415 & .296 & .468 & .404 \\
Universal-PRM-Qwen2.5-Math-7B & 690K & \underline{.534} & .624 & \textbf{.329} & .730 & .753 & .691 & .328 & .330 & .496 & .566 & .540 \\
Qwen2.5-Math-7B-PRM800k & 265K & \textbf{.586} & .613 & \underline{.301} & .708 & .768 & .727 & .362 & .404 & \underline{.500} & .594 & .559 \\
Qwen2.5-Math-PRM-7B & 860K & .531 & \textbf{.702} & \underline{.310} & .711 & .757 & .745 & .334 & .429 & \textbf{.514} & .595 & \underline{.565} \\
\midrule
\rowcolor{gray!20}
 \multicolumn{13}{c}{\textit{Reasoning Probes (ReProbes)}} \\
\midrule 
$\bigstar$ ReProbe, Attn+Logit, Self-anno (Ours) & \textbf{32K} & .529 & .594 & .260 & .735 & .779 & .779 & \underline{.394} & .404 & .461 & \underline{.618} & .559 \\
$\bigstar$ ReProbe, Attn+Logit, DeepSeek-anno (Ours) &\textbf{32K} & .465 & .616 & .243 & \underline{.740} & \underline{.802} & \underline{.786} & \textbf{.395} & .361 & .441 & \underline{.617} & .551 \\
\midrule

$\bigstar$ ReProbe, Hidden States, Self-anno (Ours) & \textbf{32K} & \underline{.558} & \underline{.673} & .264 & \textbf{.799} & \textbf{.819} & \underline{.785} & .381 & \textbf{.553} & \underline{.498} & \textbf{.667} & \textbf{.604} \\
$\bigstar$ ReProbe, Hidden States, DeepSeek-anno (Ours) &\textbf{32K} & .534 & \underline{.650} & .281 & \underline{.793} & \underline{.817}  & \textbf{.789} & .344 & .450 & .488 & .639 & \underline{.582} \\
\bottomrule
\end{tabular}
}
\caption{PR-AUC$\uparrow$ 
for detecting incorrect reasoning steps (Qwen3-8B). Best scores are shown in \textbf{bold}. Other competitive scores show clear advantages are \underline{underlined}. \#~Sample indicates the number of training samples; each sample corresponds to a reasoning trajectory with step-level labels.
}
\vspace{-0.7em}
\label{tab:step_level}


\end{table*}

\section{Experimental Setup} \label{sec:exp_setup}

\subsection{LLMs for Reasoning, ReProbe Training, Evaluation Datasets, and Baselines}

\textbf{LLMs for reasoning} used in our experiments include three state-of-the-art models: \texttt{Qwen3-8B} \citep{yang2025qwen3} in non-thinking CoT mode, \texttt{Qwen3-1.7B} and \texttt{Qwen3-32B} in native thinking mode, and \texttt{Phi-4} \cite{abdin2025phi}.


\myparagraph{Probe training settings.} We use \citet{shelmanov2025headpredictheadquestion}'s framework to train ReProbes. We conducted a vast hyperparameter search; the best training hyperparameters are detailed in \cref{appendix:training_details}.

\myparagraph{Baselines.}
We benchmark against a broad set of baselines, covering  (1) two small 1.5B PRMs; 
(2)~six large state-of-the-art 7-8B PRMs; (3) basic and state-of-the-art UQ methods implemented in the LM-Polygraph framework \citep{fadeeva-etal-2023-lm,vashurin2024benchmarking}.
More details about the baselines are presented in \cref{app:baseline}.

\noindent \textbf{Evaluation datasets} span three domains: mathematical reasoning (in-domain), planning (OOD), and general knowledge QA (OOD). We select datasets that demand non-trivial reasoning and include reasoning steps that are unambiguously verifiable, enabling reliable evaluation by both LLMs and human annotators. Test dataset details are given in \cref{appendix:test_set_details,appendix:dataset_examples}.

\subsection{Evaluation Settings}

We conduct experiments in three settings: (1)~step-level correctness assessment; (2)~best-of-$N$ TTS; and (3)~beam search TTS.

\myparagraph{(1) Step-level correctness.} For each question in the test set, we generate a CoT trace using the Qwen or Phi models and evaluate the correctness of their reasoning steps with DeepSeek-R1. To ensure accurate judgments, DeepSeek-R1 is provided with the ground-truth answer, reasoning steps, and supporting evidence (see the prompt in \cref{appendix:anno_prompt}). 
To validate the reliability of the judge, we evaluate it against (1) the human annotations from a random subset of PRM800k and (2)~a manually annotated set of 
1000 steps spanning QA, planning, and ProofNet tasks. DeepSeek-R1 achieves 95\% acc. on PRM800k and \mytexttilde90\% on other datasets (see \cref{tab:steps_acc} and \cref{appendix:human_validation} for details). Based on the judges' assessments, we compute the PR-AUC evaluation metric. 

\myparagraph{(2) Best-of-$N$} TTS setting 
assumes that 
we generate $N$ reasoning trajectories per problem ($N{=}10$ for the mathematical and QA datasets, $N{=}5$ for the planning datasets, temperature=1.0). 
The evaluation metric in this setting is the accuracy of the final solution. For GSM8K, we use an exact match against the gold-standard final answer. For other datasets, where final answers may be open-ended or structurally complex, we use DeepSeek-R1 to assign binary correctness labels (1 for correct, 0 for incorrect) based on both the problem statement and the reference solution. The grading prompt is provided in \cref{appendix:anno_prompt}.




\myparagraph{(3) Beam search} TTS setting 
assumes that we maintain a beam of $B=5$ partial reasoning trajectories at each step. From each partial trajectory, we sample $N=5$ candidate continuations, resulting in $B \cdot N$ potential candidates at each step. We assign a step-level correctness score to each candidate and retain the top-$B$ expanded trajectories with the highest scores. The correctness of the final solution is evaluated following the BoN setting, using overall accuracy as the metric.

\begin{table*}[t]

\centering

\resizebox{\linewidth}{!}{
\begin{tabular}{lc|ccc|cc|cc|ccc}
\toprule
\multirow{2}{*}{\textbf{Method}} & \multirow{2}{*}{\textbf{\# Sample}} & \multicolumn{3}{c|}{\textbf{Math (ID)}} & \multicolumn{2}{c|}{\textbf{Planning (OOD)}} & \multicolumn{2}{c|}{\textbf{QA (OOD)}} & \multicolumn{3}{c}{\textbf{Average}} \\
 &  & \textbf{MATH} & \textbf{GSM8k} & \textbf{ProofNet} & \textbf{Trips} & \textbf{Calendar} & \textbf{StrQA} & \textbf{SciQA} & \textbf{ID} & \textbf{OOD} & \textbf{Overall}\\
\midrule
\rowcolor{gray!20}
 \multicolumn{12}{c}{\textit{PRMs 750$\times$ to 810$\times$ Larger than ReProbes}} \\
\midrule
Qwen2.5-Math-7B-PRM800k               & 263K & \underline{89.8} & 80.4 & \underline{95.2} & 13.1 & \underline{45.0} & 86.7 & {91.2} & 88.5 & 59.0 & 71.6 \\
Qwen2.5-Math-PRM-7B                   & 860K & \underline{88.1} & \underline{95.4} & 93.6 & 13.1 & 41.5 & 79.2 & 83.6 & \underline{92.4} & 54.4 & 70.7 \\
\midrule
\rowcolor{gray!20}
 \multicolumn{12}{c}{\textit{Reasoning Probes (ReProbes)}} \\
\midrule

$\bigstar$ ReProbe, Attn+Logit, Self-anno (Ours)     & \textbf{32K} & \textbf{90.3} & \underline{95.4} & \underline{95.1} & \underline{15.0} & 41.5 & \underline{94.8} & \underline{96.2} & \underline{93.6} & \underline{61.9} & \underline{75.5} \\
$\bigstar$ ReProbe, Attn+Logit, DeepSeek-anno (Ours) & \textbf{32K} & 81.8 & 93.3 & 79.1 & 11.8 & \underline{42.4} & {90.7} & 90.8 & 84.7 & 58.9 & 70.0 \\
\midrule
$\bigstar$ ReProbe, Hidden States, Self-anno (Ours) & \textbf{32K} & 84.1 & \underline{97.3} & 90.6 & \textbf{15.6} & 40.5 & \underline{91.2} & \underline{92.8} & 90.7 & \underline{60.0} & \underline{73.2} \\
$\bigstar$ ReProbe, Hidden States, DeepSeek-anno (Ours) & \textbf{32K} & {86.8} & \textbf{98.8} & \textbf{95.6} & \underline{13.4} & \textbf{48.0} & \textbf{96.7} & \textbf{96.7} & \textbf{93.7} & \textbf{63.7} & \textbf{76.6} \\
\bottomrule
\end{tabular}
}
\caption{Beam search decoding accuracy across datasets (Qwen3-8B).}
\label{tab:beamsearch}

\end{table*}

\begin{table*}[t]

\centering

\resizebox{\linewidth}{!}{%

\begin{tabular}{lc|ccc|ccc|cc|ccc}
\toprule
\multirow{2}{*}{\textbf{Method}} & \multirow{2}{*}{\textbf{\# Sample}} & \multicolumn{3}{c|}{\textbf{Math (ID)}} & \multicolumn{3}{c|}{\textbf{Planning (OOD)}} & \multicolumn{2}{c|}{\textbf{QA (OOD)}} & \multicolumn{3}{c}{\textbf{Average}} \\
 &  & \textbf{MATH} & \textbf{GSM8k} & \textbf{ProofNet} & \textbf{Trips} & \textbf{Meetings} & \textbf{Calendar} & \textbf{StrQA} & \textbf{SciQA} & \textbf{ID} & \textbf{OOD} & \textbf{Overall}\\
 \midrule
 \rowcolor{gray!20}
 \multicolumn{13}{c}{\textit{Pass@N, Majority Voiting, or Larger LLM}} \\
\midrule
Qwen3-8B pass@1 (Lower Bound)            & - & 92.4 & 95.6 & 74.1 & 8.1 & 5.5 & 23.5 & 86.8 & 92.7 & 87.4 & 43.3 & 59.8 \\
Qwen3-8B pass@$N$ (Upper Bound)         & - & 99.3 & 99.2 & 97.8 & 36.2 & 16.0 & 60.5 & 98.3 & 99.3 & 98.8 & 56.1 & 72.1 \\
Qwen3-14B pass@1            & - & 93.4 & 97.6 & 76.0 & 38.7 & 5.0 & 40.5 & 91.9 & 95.6 & 89.0 & 54.3 & 67.3 \\
Majority Voting           & - & -- & \underline{97.6} & -- & -- & -- & -- & 86.6 & 92.5 & -- & -- & -- \\
\midrule
\rowcolor{gray!20}
 \multicolumn{13}{c}{\textit{PRMs \textbf{150$\times$} Larger than ReProbes}} \\
\midrule
Skywork-PRM-1.5B              & Unk & \underline{94.4} & \underline{97.6} & \underline{76.5} & 6.9 & 6.0 & 23.0 & 86.6 & 96.3 & \underline{89.5} & 43.8 & 60.9 \\
H4-Qwen2.5-PRM-1.5B-0.2      & 369K & 91.7 & 95.1 & 71.4 & \underline{15.6} & 4.5 & 22.0 & 84.6 & 94.7 & 86.1 & 44.3 & 59.9 \\
\midrule
\rowcolor{gray!20}
 \multicolumn{13}{c}{\textit{PRMs \textbf{750$\times$} to \textbf{810$\times$} Larger than ReProbes}} \\
\midrule
Math-Shepherd-PRM-7B          & 440K & 93.0 & 95.5 & 72.8 & 9.1 & 4.0 & 30.0 & 87.3 & 95.8 & 87.1 & 45.2 & 60.9 \\
RLHFlow-PRM-Deepseek-Data & 253K & 92.7 & 96.4 & 71.7 & 8.7 & 3.5 & 25.5 & 87.6 & 94.1 & 86.9 & 43.9 & 60.0 \\
RLHFlow-PRM-Mistral-Data  & 273K & 93.7 & 96.3 & 71.7 & 8.4 & 3.5 & \underline{30.0} & 87.8 & 94.1 & 87.2 & 44.8 & 60.7 \\
Universal-PRM-Qwen2.5-Math-7B  & 690K & \textbf{95.7} & 97.5 & 76.0 & 9.7 & 4.0 & 24.0 & 87.8 & \textbf{97.1} & \textbf{89.7} & 44.5 & 61.5 \\
Qwen2.5-Math-7B-PRM800k       & 263K & 92.7 & 97.3 & 74.4 & 5.9 & 6.0 & 27.5 & 87.1 & \underline{96.9} & 88.1 & 44.7 & 61.0 \\
Qwen2.5-Math-PRM-7B           & 860K & 93.7 & \textbf{97.8} & 76.0 & 7.2 & 5.5 & 26.5 & \underline{88.1} & \underline{96.9} & \underline{89.2} & 44.8 & 61.5 \\
\midrule
\rowcolor{gray!20}
 \multicolumn{13}{c}{\textit{Reasoning Probes (ReProbes)}} \\
\midrule
$\bigstar$ ReProbe, Attn+Logit, Self-anno (Ours)              & \textbf{32K} & \underline{94.4} & 97.5 & 73.6 & 9.4 & \underline{6.5} & \textbf{31.0} & \textbf{88.6} & \textbf{97.1} & 88.5 & \underline{46.5} & \underline{62.3} \\
$\bigstar$ ReProbe, Attn+Logit, DeepSeek-anno (Ours)          & \textbf{32K} & 92.7 & \textbf{97.8} & \underline{76.5} & \textbf{17.2} & \textbf{7.0} & 26.0 & \textbf{88.6} & \underline{96.9} & \underline{89.0} & \textbf{47.1} & \textbf{62.8} \\
\midrule
$\bigstar$ ReProbe, Hidden States, Self-anno (Ours) & \textbf{32K} & \underline{94.4} & 96.1 & \textbf{78.5} & 13.1 & 6.0 & 26.0 & \underline{88.1} & 95.8 & \textbf{89.7} & 45.8 & \underline{62.3} \\
$\bigstar$ ReProbe, Hidden States, DeepSeek-anno (Ours) &\textbf{32K} & 92.7 & 96.1 & 74.1 & 15.0 & 5.5 & 26.0 & \textbf{88.6} & \underline{96.9} & 87.6 & \underline{46.4} & 61.9 \\
\bottomrule
\end{tabular}

}
\caption{Best-of-$N$ decoding accuracy across datasets (Qwen3-8B). For datasets with verifiable final answers, we also provide the majority voting baseline. 
\vspace{-1.0em}
}
\label{tab:bestofn_max}

\end{table*}

\section{Main Results} \label{sec:results}

\myparagraph{Step-level correctness assessment} results for Qwen3-8B are presented in \cref{tab:step_level}, and for Phi-4 and Qwen3-1.7B/32B in native thinking mode in \cref{tab:step_level_phi4,tab:natural_reasoning,tab:natural_reasoning_qwen32b} correspondingly in \cref{appendix:additional_results}.
%
%
Unsupervised UQ methods, while providing valuable signals for detecting reasoning errors, substantially underperform other approaches.
Small PRMs show clear gains over unsupervised UQ methods in the mathematical in-domain setting, but yield limited improvements on OOD datasets. 
The best large PRMs considered in our work, such as Qwen2.5-Math-7B-PRM800k and Qwen2.5-Math-PRM-7B, substantially outperform all unsupervised UQ methods and smaller PRMs on all tasks. 

Despite using 750-810$\times$ fewer parameters than PRMs, \emph{all ReProbe variants perform on par with or even surpass the best PRMs}. For ID mathematical datasets, ReProbes substantially outperform all PRM baselines except for the two strongest Qwen2.5-Math-based PRMs. Even in these cases, ReProbe performance remains highly competitive and closely matches the best PRMs.


It is not surprising that parameter-heavy PRMs achieve slightly better results on ID mathematical datasets, as they tend to strongly overfit to this domain. In contrast, ReProbe, with far fewer parameters, avoids such domain-specific overfitting. This advantage becomes evident in the average OOD PR-AUC, where all ReProbes significantly outperform the best PRMs. For all OOD planning and QA datasets, the best PR-AUC scores are consistently achieved by ReProbes. In summary, \emph{ReProbes slightly lag behind the strongest PRMs on ID tasks but outperform them on OOD tasks}. Similar results are observed for Phi-4 and for Qwen3-1.7B/32B in the native thinking mode.


Notably, the self-supervised ReProbe achieves comparable average performance to the externally supervised variant, offering an \emph{efficient self-supervised solution for reasoning step verification} that is particularly effective in OOD settings. 

Given the poor performance of unsupervised UQ in identifying incorrect steps, we exclude it from subsequent test time scaling experiments.

\myparagraph{Test time scaling.}
Best-of-$N$ results for Qwen3-8B are presented in \cref{tab:bestofn_max}, and supplementary results  are in \cref{appendix:additional_results}. 
ReProbes achieve the best performance on all datasets except MATH, where ReProbe ranks second best. Notably, ReProbe-based TTS enables Qwen3-8B to outperform its larger sibling -- Qwen3-14B on multiple benchmarks: MATH, GSM8K, ProofNet, Meeting Planning, and ScienceQA. 
As in the step assessment setting, ReProbe exhibits strong generalization. While best PRMs (Universal-PRM-7B, Qwen2.5-MATH-7B-PRM, and Qwen2.5-Math-PRM-7B) reach parity with ReProbe on in-domain datasets MATH and GSM8K, they often fall behind on OOD tasks, such as planning and StrategyQA.

Beam search
TTS results with Qwen3-8B are reported in \cref{tab:beamsearch}. 
Across all in-domain and out-of-domain datasets, ReProbe achieves the strongest performance, outperforming state-of-the-art PRMs.

Overall, \textit{ReProbe provides stable gains in test time scaling across tasks of varying difficulty, from relatively simple datasets (MATH, GSM8K, ScienceQA) to complex planning benchmarks.}


\begin{table*}[t]
\centering
\centering

\resizebox{\linewidth}{!}{
\begin{tabular}{lc|ccc|ccc|cc|ccc}
\toprule
\multirow{2}{*}{\textbf{Method}} & \multirow{2}{*}{\textbf{\# Sample}} &
\multicolumn{3}{c|}{\textbf{Math (ID)}} &
\multicolumn{3}{c|}{\textbf{Planning (OOD)}} &
\multicolumn{2}{c|}{\textbf{QA (OOD)}} &
\multicolumn{3}{c}{\textbf{Average}} \\
 &  & \textbf{MATH} & \textbf{GSM8k} & \textbf{ProofNet}
 & \textbf{Trips} & \textbf{Meetings} & \textbf{Calendar}
 & \textbf{StrQA} & \textbf{SciQA}
 & \textbf{ID} & \textbf{OOD} & \textbf{Overall} \\
 \midrule
\rowcolor{gray!20}
 \multicolumn{13}{c}{\textit{PRMs \textbf{750$\times$} to \textbf{810$\times$} Larger than ReProbes}} \\
 
\midrule





Math-Shepherd-PRM-7B & 440K &
.049 & .248 & .234 &
.219 & .481 & .338 &
.265 & .313 &
.177 & .323 & .268 \\

RLHFlow-PRM-Deepseek-Data & 253K &
.063 & .267 & .228 &
.169 & .498 & .322 &
\textbf{.449} & \underline{.368} &
.186 & .361 & .296 \\

RLHFlow-PRM-Mistral-Data & 273K &
.047 & .184 & .164 &
.141 & .457 & .275 &
.273 & .279 &
.132 & .285 & .228 \\

Universal-PRM-Qwen2.5-Math-7B & 690K &
.175 & .331 & .299 &
.272 & .708 & .446 &
.335 & .329 &
.268 & .418 & .362 \\

Qwen2.5-Math-7B-PRM800K & 263K &
.217 & \underline{.392} & \textbf{.505} &
.270 & \underline{.716} & .587 &
.391 & .336 &
\underline{.371} & \underline{.460} & \underline{.427} \\

Qwen2.5-Math-PRM-7B & 860K &
\underline{.293} & .270 & \underline{.465} &
\underline{.328} & \textbf{.732} & \textbf{.605} &
.378 & .293 &
.343 & \underline{.467} & \underline{.421} \\

\midrule
\rowcolor{gray!20}
 \multicolumn{13}{c}{\textit{Reasoning Probes (ReProbes)}} \\
 \midrule
ReProbe, Hidden States, GPT-OSS-anno (Ours) & \textbf{10.8K} &
\textbf{.312} & \textbf{.433} & \underline{.465} &
\textbf{.334} & .505 & \underline{.604} &
\underline{.426} & \textbf{.451} &
\textbf{.403} & \textbf{.474} & \textbf{.447} \\

\bottomrule
\end{tabular}

}
\caption{PR-AUC$\uparrow$ 
for detecting incorrect reasoning steps for Qwen3-1.7B in native thinking mode.}

\label{tab:natural_reasoning}
\end{table*}

\begin{table*}[t]
\centering
\centering

\resizebox{\linewidth}{!}{
\begin{tabular}{lc|ccc|ccc|cc|ccc}
\toprule
\multirow{2}{*}{\textbf{Method}} & \multirow{2}{*}{\textbf{\# Sample}} &
\multicolumn{3}{c|}{\textbf{Math (ID)}} &
\multicolumn{3}{c|}{\textbf{Planning (OOD)}} &
\multicolumn{2}{c|}{\textbf{QA (OOD)}} &
\multicolumn{3}{c}{\textbf{Average}} \\
 &  & \textbf{MATH} & \textbf{GSM8k} & \textbf{ProofNet}
 & \textbf{Trips} & \textbf{Meetings} & \textbf{Calendar}
 & \textbf{StrQA} & \textbf{SciQA}
 & \textbf{ID} & \textbf{OOD} & \textbf{Overall} \\
 \midrule
\rowcolor{gray!20}
 \multicolumn{13}{c}{\textit{PRMs \textbf{750$\times$} to \textbf{810$\times$} Larger than ReProbes}} \\
 
\midrule

Skywork-PRM-1.5B & Unk & 
.303 & .886 & .289 &
.420 & .565 & .508 &
\textbf{.527} & .265 & .493 & .457 & .470 \\

H4-Qwen2.5-PRM-1.5B-0.2  & 369K & 
.133 & .464 & .149 &
.470 & .325 & .386 &
.129 & .221 & .249 & .306 & .285 \\

Universal-PRM-Qwen2.5-Math-7B & 690K &
.578 & \underline{.903} & .568 & 
\underline{.737} & .634 & .637 &
.189 & .303 & \underline{.683} & .500 & .569 \\

Qwen2.5-Math-7B-PRM800K & 263K &
.409 & .894 & .454 &
.573 & .591 & .594 &
.282 & \textbf{.350} & .586 & .478 & .518 \\

Qwen2.5-Math-PRM-7B & 860K &
\underline{.668} & \textbf{.930} & \underline{.646} &
.696 & \underline{.660} & \underline{.676} &
\underline{.313} & \underline{.315} & \textbf{.748} & \underline{.532} & \textbf{.613} \\

\midrule
\rowcolor{gray!20}
 \multicolumn{13}{c}{\textit{Reasoning Probes (ReProbes)}} \\
 \midrule
ReProbe, Hidden States, GPT-OSS-anno (Ours) & \textbf{10.8K} &
\textbf{.676} & .535 & \textbf{.682} &
\textbf{.752} & \textbf{.734} & \textbf{.698} &
.302 & .303 & .631 & \textbf{.558} & \underline{.585} \\

\bottomrule
\end{tabular}

}
\caption{PR-AUC$\uparrow$ 
for detecting incorrect reasoning steps for Qwen3-32B in native thinking mode.}
\vspace{-0.7em}
\label{tab:natural_reasoning_qwen32b}
\end{table*}

\begin{table*}[t]
\centering
\centering

\resizebox{0.85\linewidth}{!}{
\begin{tabular}{lc|ccc|cc|ccc}
\toprule
\multirow{2}{*}{\textbf{Method}} & \multirow{2}{*}{\textbf{\# Sample}} &
\multicolumn{3}{c|}{\textbf{Math (ID)}} &
\multicolumn{2}{c|}{\textbf{QA (OOD)}} &
\multicolumn{3}{c}{\textbf{Average}} \\
 &  & \textbf{MATH} & \textbf{GSM8k} & \textbf{ProofNet}
 & \textbf{StrQA} & \textbf{SciQA}
 & \textbf{ID} & \textbf{OOD} & \textbf{Overall} \\

 \midrule
 \rowcolor{gray!20}
 \multicolumn{10}{c}{\textit{Pass@N}} \\
\midrule
 
Qwen3-1.7B pass@1 (Lower Bound) & -- &
54.6 & 70.1 & 47.7 &
43.2 & 48.3 &
57.5 & 45.8 & 52.8
\\

Qwen3-1.7B pass@N (Upper Bound) & -- &
83.1 & 93.3 & 77.4 &
74.0 & 89.2 &
84.6 & 81.6 & 83.4
\\


 \midrule
\rowcolor{gray!20}
 \multicolumn{10}{c}{Unsupervised Uncertainty Quantification (UQ)} \\
 \midrule

MaxProb & -- &
55.6 & 84.0 & 52.9 &
50.9 & \underline{59.7} &
64.2 & \textbf{55.3} & 60.6
\\

MaxEntropy & -- &
62.2 & 83.6 & 47.1 &
49.4 & \textbf{60.0} &
\underline{64.3} & \underline{54.7} & 60.5
\\

Perplexity & -- &
55.6 & 84.0 & 52.9 &
50.9 & \underline{59.7} &
64.2 & \textbf{55.3} & 60.6
\\

\midrule
\rowcolor{gray!20}
 \multicolumn{10}{c}{\textit{PRMs \textbf{150$\times$} Larger than ReProbes}} \\
\midrule

H4-Qwen2.5-PRM-1.5B-0.2 & 369K &
63.7 & 78.3 & 52.9 &
45.4 & 47.2 &
65.0 & 46.3 & 57.5
\\

\midrule
\rowcolor{gray!20}
 \multicolumn{10}{c}{\textit{PRMs \textbf{750$\times$} to \textbf{810$\times$} Larger than ReProbes}} \\
 \midrule

Math-Shepherd-PRM-7B & 440K &
\textbf{67.3} & 84.6 & \underline{56.2} &
\textbf{54.6} & 47.2 &
\underline{69.4} & 50.9 & \underline{62.0}
\\

RLHFlow-PRM-Deepseek-Data & 253K &
\underline{64.9} & 82.8 & 49.6 &
50.0 & 45.6 &
65.8 & 47.8 & 58.6
\\

RLHFlow-PRM-Mistral-Data & 273K &
62.6 & 82.4 & 48.8 &
\underline{51.2} & 49.8 &
64.6 & 50.5 & 59.0
\\

Universal-PRM-Qwen2.5-Math-7B & 690K &
\textbf{67.3} & \underline{85.0} & \textbf{57.0} &
48.5 & 57.4 &
\textbf{69.8} & 53.0 & \textbf{63.0}
\\

Qwen2.5-Math-PRM-7B & 860K &
62.6 & \textbf{85.4} & 52.9 &
44.8 & 51.5 &
67.0 & 48.2 & 59.4
\\

\midrule
\rowcolor{gray!20}
 \multicolumn{10}{c}{\textit{Reasoning Probes (ReProbes)}} \\
 \midrule
ReProbe, Hidden States, GPT-OSS-anno (Ours) & \textbf{10.8K} &
62.2 & \underline{85.0} & \textbf{57.9} &
\underline{51.8} & 45.9 &
68.4 & 48.9 & \underline{60.6}
\\

\bottomrule
\end{tabular}

}
\vspace{-0.3em}
\caption{Best-of-$N$=10 decoding accuracy across datasets for Qwen3-1.7B in native thinking mode.}
\vspace{-0.4em}
\label{tab:bon_natural_reasoning}
\end{table*}

\begin{table*}[h]
\centering

\resizebox{\linewidth}{!}{
\begin{tabular}{lc|ccc|ccc|cc|ccc}
\toprule
\multirow{2}{*}{\textbf{Method}} & \multirow{2}{*}{\textbf{\# Sample}} 
 & \multicolumn{3}{c|}{\textbf{Math (ID)}} 
 & \multicolumn{3}{c|}{\textbf{Planning (OOD)}} 
 & \multicolumn{2}{c|}{\textbf{QA (OOD)}} 
 & \multicolumn{3}{c}{\textbf{Average}} \\
 &  & \textbf{MATH} & \textbf{GSM8k} & \textbf{ProofNet} 
 & \textbf{Trips} & \textbf{Meetings} & \textbf{Calendar} 
 & \textbf{StrQA} & \textbf{SciQA} 
 & \textbf{ID Avg.} & \textbf{OOD Avg.} & \textbf{Overall Avg.} \\
\midrule
ReProbe, Attn+Logit, Indiverse & 6K & .308 & .549 & \textbf{.205} & .626 & .687 & .685 & \textbf{.377} & .251 & .354 & .525 & .461 \\
ReProbe, Attn+Logit, Diverse & 6K  & \textbf{.409} & \textbf{.575} & .180 & \textbf{.707} & \textbf{.793} & \textbf{.792} & .271 & \textbf{.325} & \textbf{.388} & \textbf{.578} & \textbf{.507} \\
\bottomrule
\end{tabular}

}
\vspace{-0.3em}
\caption{PR-AUC of ReProbes trained on the diverse vs. homogeneous subsets of 2K questions (with three trajectories sampled per question). Increased data diversity consistently improves ReProbe’s overall performance.}
\label{tab:diversity}
\vspace{-0.7em}
\end{table*}

\myparagraph{Native thinking mode.}
A key question is whether ReProbe remains effective when reasoning is produced in the model’s native \texttt{<think>} format rather than as explicitly structured steps. To test this, we train ReProbe on Qwen3-1.7B and Qwen3-32B in native thinking mode, treating each sentence as a reasoning step and using GPT-OSS-120B to annotate step correctness conditioned on the preceding context. The step-level results for both models (\cref{tab:natural_reasoning,tab:natural_reasoning_qwen32b}) show the same overall pattern as in the structured-CoT setting: ReProbe remains competitive with or stronger than PRMs on both ID and OOD average PR-AUC. This advantage also carries over to test-time scaling: in the native-thinking Best-of-$N$ setting for Qwen3-1.7B (\cref{tab:bon_natural_reasoning}), ReProbe improves final answer selection. Overall, these results suggest that ReProbe does not depend on prompt-engineered step formatting and can also supervise more realistic free-form reasoning traces.

\begin{table}

  \centering
  \scriptsize
  \setlength{\tabcolsep}{6pt}
  \resizebox{.49\textwidth}{!}{
  \begin{tabular}{l ccc}
    \toprule
    \textbf{Method} & \textbf{MATH} & \textbf{GSM8k} & \textbf{ProofNet} \\
    \midrule
    PRM1 (Qwen2.5-Math-7B-PRM800k)  & .586 & .613 & .301 \\
    PRM2 (Qwen2.5-Math-7B)          & .531 & .702 & .310 \\
    ReProbe, Attn+Logit, DeepSeek-anno      & .529 & .594 & .260 \\
    \midrule
    ReProbe + PRM1    & \textbf{.613} & .674 & .318 \\
    ReProbe + PRM2    & .573 & \textbf{.710} & \textbf{.327} \\
    \bottomrule
  \end{tabular}
  }
\caption{Synergy of PRMs with ReProbes. Step-level PR-AUC, Qwen3-8B.
\vspace{-1.5em}
}
  \label{tab:uhead_plus_prm}
\end{table}

\section{Analysis} \label{sec:analysis}


\myparagraph{Synergy of PRMs and ReProbes.}
PRMs and ReProbes capture complementary aspects of reasoning quality: PRMs rely on their own knowledge and textual cues from the generated rationale, whereas probes exploit internal signals of the LLM that encode model confidence. To explore whether these two perspectives can be synergistic, we train a logistic regression model that takes both the PRM score and the ReProbe score as input to predict step-level correctness labels. The model is trained on a random subset of 200 questions from the training set. \cref{tab:uhead_plus_prm} reports the step-level PR-AUC, showing that this combination yields additional improvements. This finding suggests that integrating confidence-based signal and the PRM external knowledge is a promising direction for future work.

\begin{figure}[t]
  \centering
  \includegraphics[width=\columnwidth]{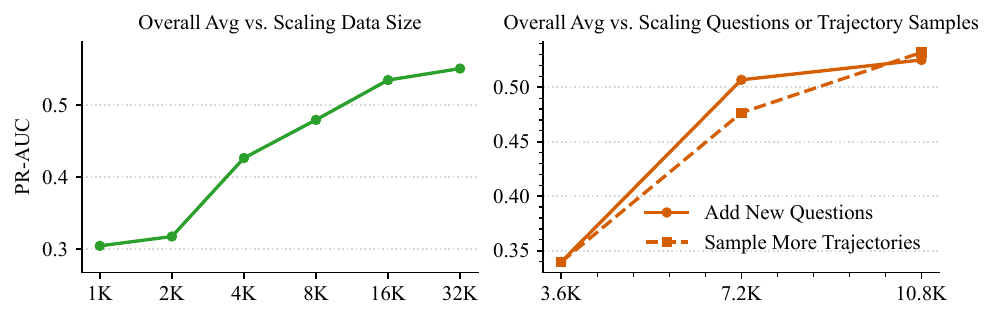}
  \caption{Left: PR-AUC of ReProbes with increasing training set size (x-axis). Right: scaling training data either by adding new unique questions or by sampling additional trajectories. Average PR-AUC across all datasets is reported. The results for each individual dataset are in \cref{fig:log-scaling-alltasks}, \cref{appendix:additional_results}.}
  \label{fig:log-scaling-avg}
  \vspace{-0.4cm}
\end{figure}

\myparagraph{Training data quantity and diversity.} We investigate how data quantity and diversity influence ReProbe performance. \cref{fig:log-scaling-avg} (left) shows that larger training sets, more questions, and sampled reasoning trajectories consistently improve ReProbe performance on both ID and OOD tasks. In practice, however, high-quality questions are often scarce, making it difficult to scale by prompt expansion alone. Alternatively, it is possible to sample multiple reasoning trajectories per question. \cref{fig:log-scaling-avg} (right) compares scaling strategies, demonstrating that trajectory sampling is an effective way to boost ReProbe performance. To study the impact of data diversity, we train ReProbes on two 2K-question subsets of training data: one highly homogeneous and one highly diverse. For similarity assessment, questions are embedded using Qwen3-Embedding-8B \citep{qwen3embedding}. The homogeneous subset is constructed by selecting the 2K nearest neighbors to the median embedding, whereas the diverse subset is obtained via farthest-first traversal \citep{GONZALEZ1985293}. \cref{tab:diversity} shows that increased data diversity leads to consistently improved ReProbe performance.

\myparagraph{Architecture.} 
We compare three architectures for ReProbe: \emph{(1) ReProbe (step-level)}: leverages a transformer encoder and predicts step-level correctness; \emph{(2) Linear Probe (token-level
)}: replacing transformer-based ReProbe with a linear layer to predict token-level labels (all tokens of incorrect step should be labeled incorrect); and \emph{(3) ReProbe (token-level)}: uses a transformer encoder and predicts token-level labels. All variants use the same training setup (hidden states as features, DeepSeek-anno).
\cref{tab:architectural_ablation} shows that step-level ReProbe outperforms both token-level variants. 


\begin{figure}[t]
  \centering
  \includegraphics[width=\columnwidth]{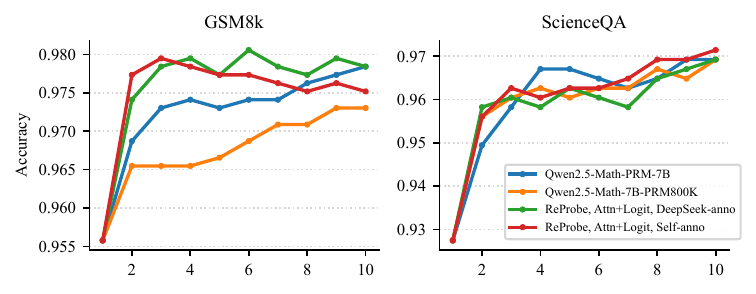}
  \caption{Best-of-$N$ accuracy on GSM8k and ScienceQA for different $N$ values (Qwen3-8B).}
  \label{fig:abl-bon}
  \vspace{-0.7em}
\end{figure}


\myparagraph{Impact of reasoning length} on the performance of ReProbes and PRMs is investigated in  \cref{appendix:analysis_performance_by_length}. We show that the performance of both PRMs and ReProbes degrades with increasing reasoning length, but only slightly.


\myparagraph{Scaling the number of samples $N$.} \Cref{fig:abl-bon} analyzes best-of-$N$ performance on GSM8k and ScienceQA for $1 \le N \le 10$. As expected, accuracy improves with increasing $N$ across methods on both datasets. On GSM8k, both ReProbe variants consistently outperform PRMs; on ScienceQA all methods exhibit comparable performance.

\myparagraph{ReProbe efficiency.} The theoretical analysis of the efficiency of ReProbe compared to PRMs is presented in \cref{appendix:reprobe_efficiency}. We also empirically illustrate the runtime efficiency of ReProbe in \cref{tab:comp_efficiency}. ReProbe in current implementation achieves a 2.6$\times$--25$\times$ speedup over state-of-the-art PRMs.

\section{Related Work}

\textbf{PRMs.} Research in PRMs has advanced by scaling and refining step-level annotations: from manual labeling \citep{uesato2022solving,lightman2023letsverifystepstep}, to MC-based automatic labeling \citep{wang-etal-2024-math,luo2024improvemathematicalreasoninglanguage}, and consensus methods combining LLM-as-a-Judge and MC estimation \citep{zhang2025lessonsdevelopingprocessreward,zhao2025genprm}. Generative PRMs further extend step-level judgment with long CoT \citep{xiong2025stepwiserstepwisegenerativejudges,zhao2025genprm}.
Our approach differs from PRMs in that ReProbe is not a language model and operates directly on internal LLM states rather than textual inputs.


\myparagraph{Uncertainty quantification} methods recently have been employed for test-time scaling and improving reasoning performance \cite{mo2024tree,yin-etal-2024-reasoning,zhang-etal-2025-entropy,fu2025deepthinkconfidence,yan2025mur,kang2025scalable}. However, so far, only unsupervised UQ methods have been used as step scorers. We propose a training-based probe and show that it achieves much better performance than unsupervised UQ.

\myparagraph{Formal verification} has recently been used to verify LLM reasoning steps \citep{zhou2024donttrustverify,hu2025stepproofstepbystepverificationnatural,liu2025safeenhancingmathematicalreasoning,zhou2025stepwiseformalverificationllmbased}. However, these methods often require specialized autoformalization data for training and are limited to narrow domains (e.g., math proofs). ReProbes are much more general; our experiments show that they can generalize to OOD.

\section{Conclusion}





We introduced ReProbe, a lightweight step-level verifier that uses an LNN’s internal states to guide reasoning. ReProbe can be trained fully self-supervised, without human labels, verifiable final answers, or Monte-Carlo rollouts. Across math, planning, and QA, it achieves strong in- and out-of-domain performance and is competitive with, or better than, far larger PRMs, making it a practical component for resource-efficient reasoning systems. 
Beyond replacing PRMs, ReProbe also complements them: combining PRM and ReProbe scores consistently improves performance, suggesting that the two capture different aspects of reasoning quality. This points to a promising path toward hybrid verifiers that combine introspective confidence with process rewards. More broadly, our results suggest a path toward more efficient test-time scaling for reasoning in LLMs.


\section*{Limitations}
\cref{sec:results} shows that ReProbe performance increases with larger training data and benefits from data diversity. Curves in \cref{fig:log-scaling-avg} and \cref{fig:log-scaling-alltasks}, although show diminishing marginal gains, do not seem to reach the top for tasks like StrategyQA. Therefore, it seems possible to further unleash the potential of ReProbe with further data scaling -- sampling more reasoning trajectories per question or adding new questions beyond the PRM800K training set. In this work, we do not involve questions outside the PRM800K training set to establish a fair comparison with PRMs trained on data derived from this set (e.g., Qwen2.5-Math-7B-PRM800K). Due to a limited budget for annotation (annotation of 32K reasoning trajectories with DeepSeek-R1 cost ${>}2000$ USD), we also do not annotate more reasoning trajectories per question. We leave this promising scaling to future work, including the integration of diverse, high-quality questions outside the math domain.

ReProbes need to be trained on the internal states of the target LLMs they supervise. Therefore, contrary to PRMs, they cannot be directly applied to another LLM since they depend on model-specific internal states. 
However, ReProbes are highly efficient in parameter size, training data, and training compute, making their training relatively inexpensive. Moreover, once trained, ReProbes can significantly reduce inference costs compared to PRMs. In practice, applications may focus on a small number of widely used models. If ReProbes for these models are shared online (e.g., on HuggingFace), practitioners could readily download and apply them without training from scratch. Future work may also investigate fine-tuning ReProbes to adapt to customized or fine-tuned versions of target LLMs.


\section*{Ethics Statement}

We use publicly available datasets (MATH, GSM8K, ProofNet, ScienceQA, and StrategyQA), which have no data privacy issues. All artifacts we use are under licenses allowing research usage. Human annotations were conducted by the authors of this paper. We do not identify any other ethical risks associated with this study.

\myparagraph{Reproducibility.} We fully open-source our trained ReProbes, code, prompts, human annotations, and processed datasets to ensure full reproducibility. For all training, evaluation, and sampling, we fix random seeds to 1 or 42 (as specified in the codebase). One major challenge in reproducing the exact numbers in our tables from scratch is the use of API-based DeepSeek-R1. API-based LLMs are known to be inherently non-deterministic even when fixing prompts and temperature. To address this, we provide all DeepSeek-R1 annotations used in training and evaluation, allowing others to faithfully reproduce our results. If reproducing from scratch, our codebase also guaranties the reproduction of similar trends and observations, even if there are slight differences in exact numbers. 

\section*{Acknowledgements}
We sincerely thank the anonymous reviewers for their valuable feedback and suggestions that helped to improve this paper.
The work was supported by the Swiss AI initiative ({\url{https://www.swiss-ai.org/compute-grants}) through a grant from the Swiss National Supercomputing Centre (CSCS) under project ID a0142 on Alps.

\bibliography{bib/bibliography}

\newpage

\appendix

\clearpage
\section{Additional Experimental Results} 
\label{appendix:additional_results}

This section provides supplementary empirical results that extend the main findings of the paper. In particular, we report results for additional target models and ablation studies that further analyze the robustness, scaling behavior, and architectural properties of ReProbe across settings.

\subsection{Results with Phi-4} \label{appendix:results_with_phi4}

To verify that our framework works for models of different sizes, families, and post-training approaches, we conduct step-level correctness and BoN evaluation on a Phi-4 ReProbe trained on Qwen3-8B annotated training data. The step-level correctness prediction and BoN results are presented in \cref{tab:step_level_phi4} and \cref{tab:bestofn_phi4} correspondingly. 
On step-level correctness, ReProbe achieves the best performance in Meeting and Calendar Planning, and the best average performance on OOD tasks. On StrategyQA and overall average, it ranks second. 

For BoN, the Qwen3-8B–annotated ReProbe outperforms two strong Qwen2.5-Math PRMs on MATH and matches the strongest PRMs on GSM8K. On ProofNet and StrategyQA, it also outperforms some much larger PRMs.


\subsection{Ablation Study on Data Quantity and ReProbe Architectures} \label{appendix:analysis_additional_result}

\cref{tab:architectural_ablation} shows that the linear probe and token-level ReProbe are worse than the proposed step-level ReProbe. \cref{fig:log-scaling-alltasks} shows the per-dataset performance improvement by scaling the data quantity.


\begin{table*}[h]
\centering

\resizebox{\linewidth}{!}{
\begin{tabular}{lc|ccc|ccc|cc|ccc}
\toprule
\multirow{2}{*}{\textbf{Method}} & \multirow{2}{*}{\textbf{\# Sample}} 
 & \multicolumn{3}{c|}{\textbf{Math (ID)}} 
 & \multicolumn{3}{c|}{\textbf{Planning (OOD)}} 
 & \multicolumn{2}{c|}{\textbf{QA (OOD)}} 
 & \multicolumn{3}{c}{\textbf{Average}} \\
 &  & \textbf{MATH} & \textbf{GSM8k} & \textbf{ProofNet} 
 & \textbf{Trips} & \textbf{Meetings} & \textbf{Calendar} 
 & \textbf{StrQA} & \textbf{SciQA} 
 & \textbf{ID Avg.} & \textbf{OOD Avg.} & \textbf{Overall} \\
\midrule
\rowcolor{gray!20}
 \multicolumn{13}{c}{\textit{Unsupervised Uncertainty Quantification (UQ)}} \\
\midrule
Random & - & .106 & .038 & .082 & .552 & .463 & .324 & .172 & .086 & .075 & .319 & .228 \\
MaxProb & - & .127 & .084 & .123 & .618 & .548 & .380 & .252 & .158 & .111 & .391 & .286 \\
MaxEntropy & - & .112 & .079 & .107 & .585 & .533 & .362 & .248 & .135 & .099 & .373 & .270 \\
Perplexity & - & .117 & .066 & .099 & .557 & .508 & .323 & .228 & .143 & .094 & .352 & .255 \\
\midrule
\rowcolor{gray!20}
 \multicolumn{13}{c}{\textit{PRMs 150$\times$ Larger than ReProbes}} \\
\midrule
Skywork-PRM-1.5B & Unk & .219 & .181 & .185 & .408 & .467 & .327 & .237 & \underline{.415} & .195 & .371 & .305 \\
H4-Qwen2.5-PRM-1.5B-0.2  & 369K & .174 & .061 & .105 &  .534 & .476 & .434 & .212 & .116 & .113 & .354 & .264 \\
\midrule
\rowcolor{gray!20}
 \multicolumn{13}{c}{\textit{PRMs 750$\times$ to 810$\times$ Larger than ReProbes}} \\
\midrule
Math-Shepherd-PRM-7B & 440K & .248 & .188 & .188 & .747 & .584 & .489 & .249 & .327 & .208 & .479 & .378 \\
RLHFlow-PRM-Deepseek-8B & 253K & .200 & .263 & .109 & .558 & .455 & .343 & .315 & \textbf{.440} & .191 & .422 & .335 \\
RLHFlow-PRM-Mistral-8B & 273K & .141 & .195 & .093 & .462 & .424 & .309 & .261 & .311 & .143 & .353 & .274 \\
Universal-PRM-Qwen2.5-Math-7B & 690K & \textbf{.485} & .213 & \textbf{.263} & .741 & .559 & .497 & .321 & .252 & .320 & .474 & .416 \\
Qwen2.5-Math-7B-PRM800k & 265K & \underline{.474} & \textbf{.406} & .238 & \textbf{.825} & \underline{.599} & \underline{.568} & \textbf{.355} & .329 & \textbf{.373} & \underline{.535} & \textbf{.474} \\
Qwen2.5-Math-PRM-7B & 860K & .427 & \underline{.377} & \underline{.240} & \underline{.791} & .594 & .555 & .333 & .310 & \underline{.348} & .517 &  \underline{.453} \\
\midrule
\rowcolor{gray!20}
 \multicolumn{13}{c}{\textit{Reasoning Probes (ReProbes)}} \\
\midrule
$\bigstar$ ReProbe, Attn+Logit, Qwen3-8B-anno (Ours) & \textbf{32K} & .404 & .340 & .155 & .756 & \textbf{.646} & \textbf{.592} & \underline{.347} & .347 & .300 & \textbf{.538} & \underline{.448} \\
\bottomrule
\end{tabular}

}
\caption{PR-AUC for detecting incorrect reasoning steps for Phi-4. Best scores are shown in \textbf{bold}, and other competitive scores are \underline{underlined}. \#~Sample indicates the number of training samples, where each sample corresponds to a reasoning trajectory with step-level annotations. $\ddagger$~Qwen2.5-Math-PRM-7B’s training data is filtered from an 860K-sample dataset; the exact size after filtering is not specified in their paper.}
\label{tab:step_level_phi4}
\end{table*}

\begin{table*}[h]

\centering

\resizebox{0.7\linewidth}{!}{
\begin{tabular}{lc|ccc|cc}
\toprule
\multirow{2}{*}{\textbf{Method}} & \multirow{2}{*}{\textbf{\# Sample}} & \multicolumn{3}{c|}{\textbf{Math (In-domain)}} & \multicolumn{2}{c}{\textbf{Reasoning QA}} \\
 &  & \textbf{MATH} & \textbf{GSM8k} & \textbf{ProofNet} & \textbf{StrQA} & \textbf{SciQA} \\
\midrule
 \rowcolor{gray!20}
 \multicolumn{7}{c}{\textit{Pass@N and Majority Voting}} \\
\midrule
Phi-4 pass@1 (Lower Bound)       & - & 90.5 & 97.9 & 86.8 & 93.0 & 95.7 \\
Phi-4 pass@$N$ (Upper Bound)     & - & 98.2 & 100. & 97.2 & 99.1 & 99.8 \\
Majority Voting                  & - &  --  & \textbf{100.} &  --  & 93.2 & 94.6 \\
\midrule
\rowcolor{gray!20}
 \multicolumn{7}{c}{\textit{PRMs 150$\times$ Larger than ReProbes}} \\
\midrule
Skywork-PRM-1.5B               & Unk & \underline{95.9} & 99.5 & \textbf{95.7} & 92.5 & 96.6 \\
H4-Qwen2.5-PRM-1.5B-0.2       & 369K & 93.5 & \textbf{100.} & 89.2 & 91.8 & 96.4 \\
\midrule
\rowcolor{gray!20}
 \multicolumn{7}{c}{\textit{PRMs 750$\times$ to 810$\times$ Larger than ReProbes}} \\
\midrule
Math-Shepherd-PRM-7B          & 440K & 92.9 & 98.4 & \underline{95.2} & 93.2 & 96.6 \\
RLHFlow-PRM-Deepseek-Data     & 253K & 94.1 & 98.9 & 89.6 & \textbf{94.8} & \textbf{98.0} \\
RLHFlow-PRM-Mistral-Data      & 273K & 92.3 & 98.4 & 90.3 & \underline{94.1} & 97.3 \\
Universal-PRM-Qwen2.5-Math-7B               & 690K & \textbf{96.4} & \textbf{100.} & 93.5 & 92.5 & 96.8 \\
Qwen2.5-Math-7B-PRM800k       & 263K & 93.5 & \textbf{100.} & 92.4 & 93.9 & \underline{97.5} \\
Qwen2.5-Math-PRM-7B & 860K & 93.5 & \textbf{100.} & 94.4 & \underline{94.1} & \underline{97.5}\\
\midrule
\rowcolor{gray!20}
 \multicolumn{7}{c}{\textit{Reasoning Probes (ReProbes)}} \\
\midrule
$\bigstar$ ReProbe, Attn+Logit, Qwen3-8B-anno (Ours)              & \textbf{32K} & \underline{95.9} & \textbf{100.} & 93.1 & 93.0 & 95.7 \\
\bottomrule
\end{tabular}
}
\caption{Best-of-$N$ decoding accuracy across datasets for Phi-4 model. For datasets with verifiable final answer, we also provide Majority Voting. 
}
\label{tab:bestofn_phi4}
\end{table*}

\begin{table*}[t]
\centering

\resizebox{\linewidth}{!}{
\begin{tabular}{lc|ccc|ccc|cc|ccc}
\toprule
\multirow{2}{*}{\textbf{Method}} & \multirow{2}{*}{\textbf{\# Sample}} & \multicolumn{3}{c|}{\textbf{Math (ID)}} & \multicolumn{3}{c|}{\textbf{Planning (OOD)}} & \multicolumn{2}{c|}{\textbf{QA (OOD)}} & \multicolumn{3}{c}{\textbf{Average}} \\
 &  & \textbf{MATH} & \textbf{GSM8k} & \textbf{ProofNet} & \textbf{Trips} & \textbf{Meetings} & \textbf{Calendar} & \textbf{StrQA} & \textbf{SciQA} & \textbf{ID} & \textbf{OOD} & \textbf{Overall}\\
\midrule
ReProbe, Step-Level (Proposed) & \textbf{32K} 
& \underline{.534} & \underline{.650} & \underline{.281} 
& \underline{.793} & \underline{.817} & \underline{.789} 
& \underline{.344} & \underline{.450} 
& \underline{.488} & \underline{.639} & \underline{.582} \\

ReProbe, Token-Level & \textbf{32K} 
& \textcolor{deepred}{--.046} & \textcolor{deepred}{--.049} & \textcolor{deepred}{--.072} 
& \textcolor{deepred}{--.129} & \textcolor{deepred}{--.070} & \textcolor{deepred}{--.017} 
& \textcolor{deepred}{--.021} & \textcolor{deepred}{--.075} 
& \textcolor{deepred}{--.055} & \textcolor{deepred}{--.063} & \textcolor{deepred}{--.060} \\

Linear Probe, Token-Level & \textbf{32K} 
& \textcolor{deepred}{--.112} & \textcolor{deepred}{--.036} & \textcolor{deepred}{--.063} 
& \textcolor{deepred}{--.157} & \textcolor{deepred}{--.224} & \textcolor{deepred}{--.422}  
& \textcolor{deepgreen}{+.011} & \textcolor{deepgreen}{+.009} 
& \textcolor{deepred}{--.070} & \textcolor{deepred}{--.157} & \textcolor{deepred}{--.124} \\
\bottomrule
\end{tabular}
}
\caption{Replacing the proposed ReProbe with other architectures. Change in PR-AUC relative to the original step-level ReProbe.}

\label{tab:architectural_ablation}
\end{table*}

\begin{figure*}[t]
  \centering
  \includegraphics[width=\linewidth]{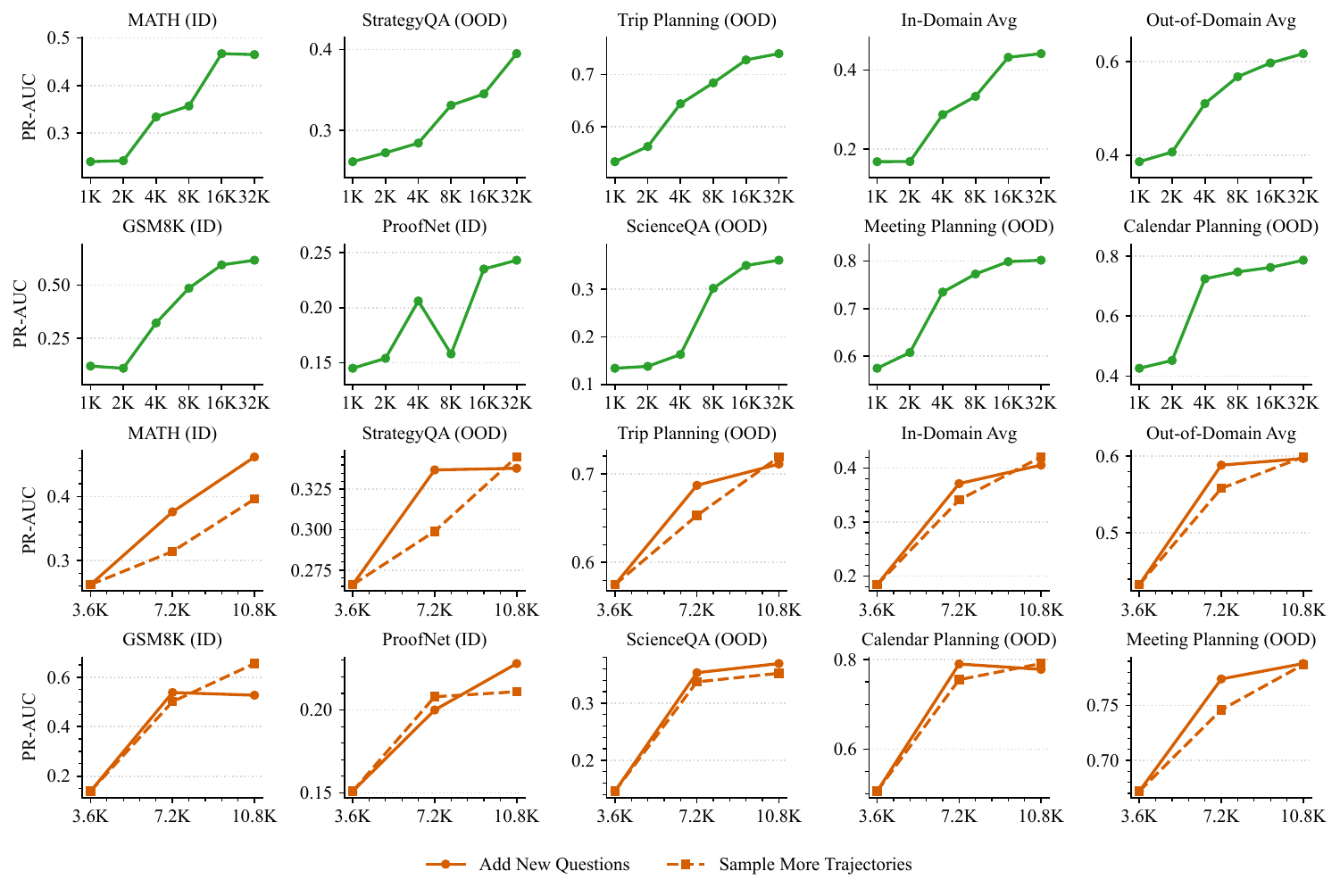}
  \caption{Top 2 rows: PR-AUC of ReProbes with increasing training set size (x-axis). Bottom 2 rows: scaling training data either by adding new unique questions or by sampling additional trajectories. 
  }
  \label{fig:log-scaling-alltasks}
\end{figure*}

\subsection{Impact of Reasoning Length on ReProbe Performance}
\label{appendix:analysis_performance_by_length}

\begin{figure}[t]
  \centering
  \includegraphics[width=1.0\linewidth]{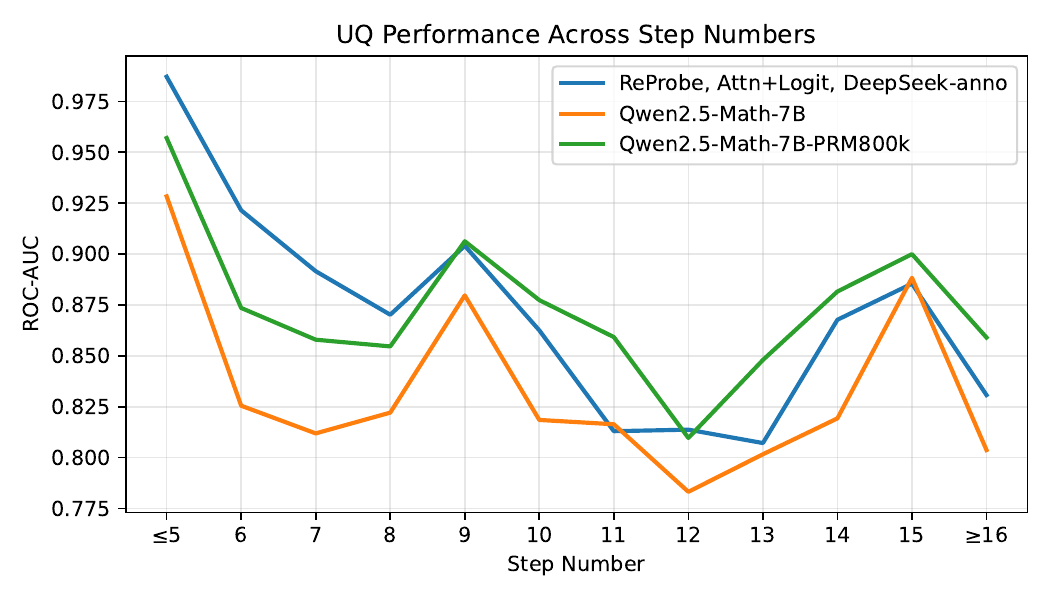}
  \caption{ROC-AUC on the step-level MATH benchmark using Qwen3-8B, evaluated across bins of reasoning chain length (number of reasoning steps), for the proposed ReProbe and two PRM models.}
  \label{fig:performance_by_length}
\end{figure}

Figure~\ref{fig:performance_by_length} compares the performance of ReProbe with two selected PRM models across different bins of total reasoning steps. The results indicate that ReProbe consistently outperforms the PRM baselines on shorter generations (i.e., $\leq 8$ reasoning steps), while achieving comparable performance as the reasoning chains become longer.

\subsection{Token-Level Feature Importance Analysis}

To better understand where the uncertainty signal captured by ReProbe arises, we conducted a token-level feature importance analysis.

For each token in a reasoning step, we replace that token’s feature vector with the mean hidden feature vector over the entire generation and measure the absolute change in the predicted uncertainty score. This change is treated as the token's importance.

We find that the signal is not uniformly distributed across tokens. Instead, it is often concentrated near step boundaries. In approximately 40\% of reasoning steps the most important token is the final token, while in 12.5\% of steps it is the first token.

Moreover, the most influential tokens frequently correspond to sentence boundary markers, especially punctuation tokens. The token ``.'' is the most important token in approximately {27\% of reasoning steps}. Interestingly, the token {``Wait''} appears as the most important token in about {2.5\% of reasoning steps}, consistent with its role as a marker of self-correction or uncertainty during chain-of-thought reasoning.

Figure~\ref{fig:token_position_importance} shows the distribution of the relative position of the most important token within each reasoning step. The results indicate that the uncertainty signal often appears toward the {end of reasoning steps}, suggesting that the model's internal confidence is updated when a step is completed.

\begin{figure}[h]
\centering
\includegraphics[width=0.95\linewidth]{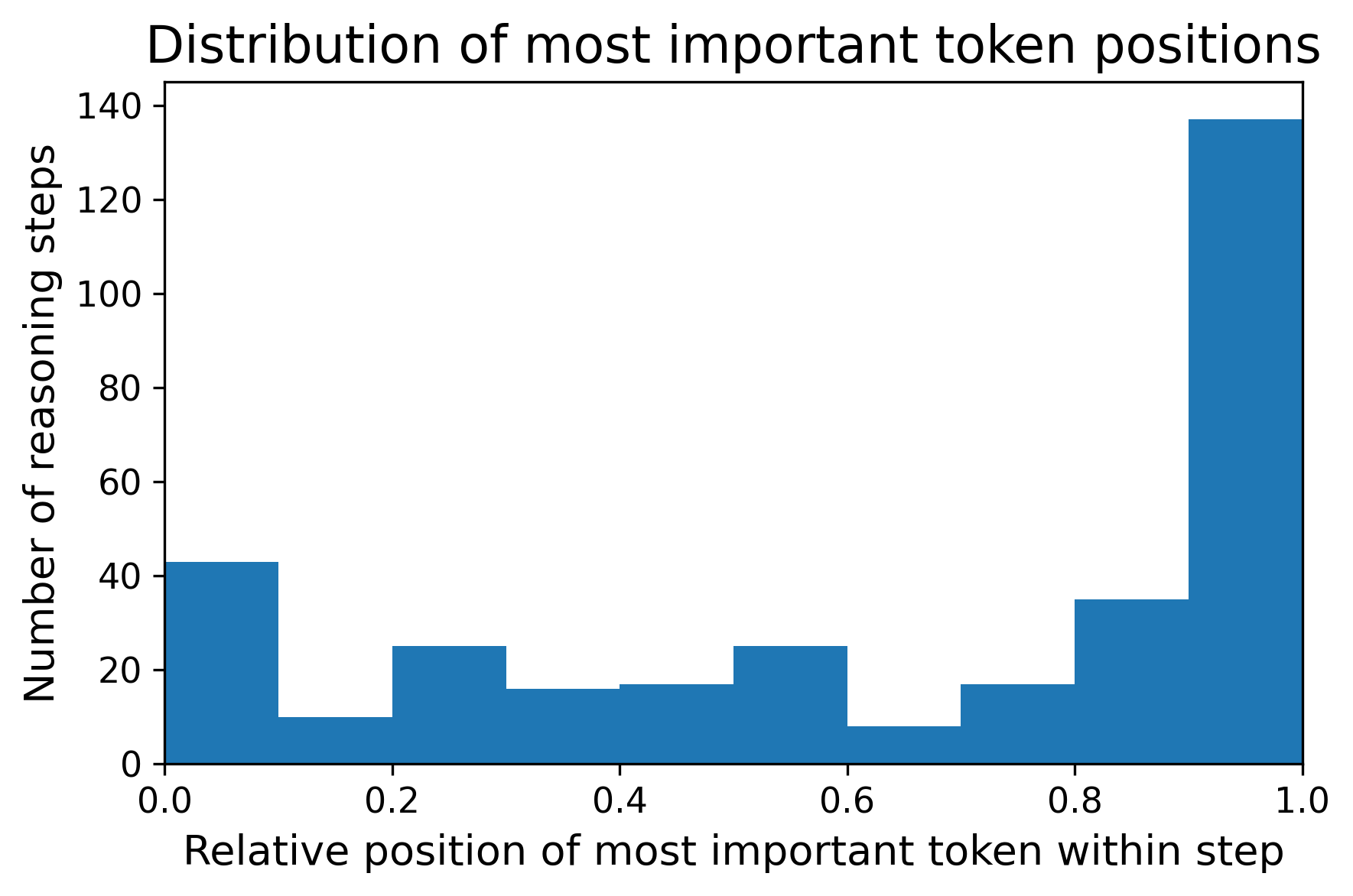}
\caption{Distribution of the relative position of the most important token within each reasoning step.}
\label{fig:token_position_importance}
\end{figure}

\subsection{Cascaded ReProbe--PRM Verification}

Table~5 in the main paper suggests that ReProbe and PRMs capture {complementary signals}: PRMs rely primarily on external knowledge and textual cues, while ReProbe leverages internal model confidence signals.

This motivates a {cascaded verification strategy}, where ReProbe acts as a lightweight first-pass filter and a computationally expensive PRM is invoked only for uncertain reasoning steps.

We evaluate a simple hybrid strategy:

\begin{enumerate}
\item Apply ReProbe to each reasoning step.
\item If the predicted uncertainty score exceeds a threshold $t$, the step is passed to a PRM.
\item Otherwise, the ReProbe prediction is used directly.
\end{enumerate}

By tuning the threshold $t$, we can trade off verification cost and performance.

\begin{table}[h]
    \centering
    \small
    \resizebox{0.95\linewidth}{!}{
    \begin{tabular}{lcc}
    \toprule
    Method & Math & ProofNet \\
    \midrule
    ReProbe, Hidden States, DeepSeek-anno  & .514 & .260 \\
    Qwen2.5-Math-7B-PRM800k & .585 & \underline{.301} \\
    \midrule
    Hybrid cascade ($t=0.1$) & \textbf{.601} & \textbf{.303} \\
    Hybrid cascade ($t=0.2$) & \underline{.590} & .273 \\
    Hybrid cascade ($t=0.3$) & .585 & .234 \\
    \bottomrule
    \end{tabular}
    }
    \caption{Hybrid cascaded verifier combining ReProbe and PRM. We report PR-AUC$\uparrow$ on 2 datasets: Math and ProofNet.}
    \label{tab:hybrid_verifier}
\end{table}

Importantly, the cascaded strategy can {match the PRM's in-domain performance while significantly reducing the number of PRM evaluations}. In our experiments, we skip approximately {56\% of PRM evaluations on MATH} (correspondning to $t=0.3$) and {31\% on ProofNet} ($t=0.1$) while maintaining comparable higher or equal (\cref{tab:hybrid_verifier}).

These results suggest that cascaded verifiers combining lightweight introspective probes with heavier reward models provide a promising direction for {efficient reasoning verification}.

\subsection{Computational Efficiency} \label{appendix:computation_efficiency}

To assess computational efficiency, we benchmarked the runtime of ReProbe and several PRM models on 500 test samples from the MATH dataset. The runtime excludes LLM generation and step-extraction overhead. For ReProbe, we measure only the feature-extraction stage together with the forward pass of the ReProbe classifier; for PRMs, we measure solely the inference time of each model. PRM runtimes are based on the official implementations provided in their respective Hugging Face repositories, while ReProbe uses our own implementation of feature extraction and inference. All evaluations were performed with a batch size of 1. \cref{tab:comp_efficiency} presents the results.

The results show that ReProbe is by far the fastest method, which is expected given its lightweight architecture and small parameter footprint. In contrast, PRM models incur substantially higher inference costs, often by one to two orders of magnitude. Minor non-monotonicities in runtime across PRMs (e.g., smaller models being slower than larger ones) are attributable to differences in engineering and implementation rather than model size. A theoretical analysis of time and memory complexity can be found in \cref{appendix:reprobe_efficiency}.

\begin{table}[h]
\centering
\tiny

\begin{tabular}{l c}
\toprule
\textbf{Method} & \textbf{Runtime} \\
\midrule
\rowcolor{gray!20}
 \multicolumn{2}{c}{\textit{PRMs \textbf{150$\times$} Larger than ReProbes}} \\
\midrule
  Skywork-PRM-1.5B & \underline{17 s} \\
  H4-Qwen2.5-PRM-1.5B-0.2 & 8 min 20 s \\
\midrule
\rowcolor{gray!20}
 \multicolumn{2}{c}{\textit{PRMs \textbf{750$\times$} to \textbf{810$\times$} Larger than ReProbes}} \\
\midrule
  Qwen2.5-Math-7B-PRM800k & 37 s \\
  Qwen2.5-Math-PRM-7B & 34 s \\
  Math-Shepherd-PRM-7B & 5 min 51 s \\
  RLHFlow-PRM-Mistral-Data & 5 min 19 s \\
  RLHFlow-PRM-Deepseek-Data & 5 min 33 s \\
  Universal-PRM-Qwen2.5-Math-7B & 4 min 22 s \\
\midrule
\rowcolor{gray!20}
 \multicolumn{2}{c}{\textit{Reasoning Probes (ReProbes)}} \\
\midrule 
 $\bigstar$ ReProbe, Hidden States, Self-anno (Ours) & \underline{14 s} \\
$\bigstar$ ReProbe, Hidden States, DeepSeek-anno (Ours) & \textbf{13 s} \\
\bottomrule
\end{tabular}
\caption{Runtime comparison of ReProbe and PRMs on 500 MATH test samples (batch size = 1).}
\label{tab:comp_efficiency}
\end{table}



\section{ReProbe Training Details} 
\label{appendix:training_reprobe_details}

This section summarizes the implementation details required to reproduce ReProbe training. We describe how the training data is annotated, list the main training hyperparameters, and report the computational and monetary cost of training.

\subsection{Training Data Annotation Details} \label{appendix:training_data_anno}

In our experiments, we employ two LLM judges for the annotation of ReProbe training data. For DeepSeek-R1, we follow the officially recommended inference setting, using a temperature of 0.6. Similarly, when using Qwen3-8B as a training data annotator, we also follow the recommended inference hyperparameter, using a temperature of 0.7, a top\_k of 20, and a top\_p of 0.95. We access Qwen3-8B through vLLM local deployment and access DeepSeek-R1 through the DeepSeek API. The annotation prompts are detailed in \cref{fig:anno_prompt_step_level}. 
For annotation efficiency, we restrict the generation of Qwen3 8B to $256$ tokens during training.

Qwen3, with native thinking enabled, generates long CoTs enclosed within \texttt{<think>} tags, without explicit step separations. To train ReProbe for native reasoning, we modify the experimental setup from 
in two ways: (1) since the native thinking trajectory is much longer than the formatted CoT, we replace the DeepSeek-R1 API with the locally deployed \texttt{GPT-OSS-120B} for step-level correctness annotation to reduce API costs. This model outperforms the January version of DeepSeek-R1 on multiple benchmarks (e.g., GPQA Diamond, AIME 2024). (2) Without format-based step separations, we treat each sentence as a step and prompt \texttt{GPT-OSS-120B} to identify errors conditioned on all previous sentences. 
For the native thinking mode, during training, we restrict the generation length to $1024$ tokens.

\subsection{Training Hyperparameter Details} \label{appendix:training_details}
For all experiments on \texttt{Phi-4} and \texttt{Qwen3-8B}, we use the same set of hyperparameters. We use a learning rate of 5e-4, a batch size of 128, and a positive class weight of 3. For ReProbes, we use one layer of the transformer encoder with a hidden size of 512 and 16 attention heads. All training continues for 5 epochs. We use 10\% of the training data and a 200-sample set of GSM8K, ScienceQA, and StrategyQA for validation and best checkpoint selection. 


\subsection{Training Cost} \label{appendix:computation_for_training}

All experiments are carried out on cluster nodes with 4 GH200 GPUs and another cluster node with 2 H100 GPUs. Training one ReProbe on 32K datapoints with Hidden States as features takes \mytexttilde4 GH200 GPU hours, while the variant with Attn+Logit as features takes \mytexttilde32 GH200 GPU hours. The training of the Attn+Logit ReProbe is slower since extracting attention maps prohibits the use of efficient attention implementations (e.g., Flash Attention).

The total annotation cost for the training dataset (32K samples drawn from PRM800k) was approximately \$200 per model. 
In addition, around \$300 was spent on evaluation across eight datasets (three for math, three for planning, and two for QA), including the annotation of both step-level correctness and complete reasoning chains for the evaluated decoding strategies.

\section{Additional Details of Evaluation Setup}

This section provides additional details of the evaluation setup used throughout the paper. We describe the baselines, test datasets, and sampling protocol in greater detail to facilitate reproducibility and clarify the scope of our comparisons.

\subsection{Baselines}
\label{app:baseline}

We benchmark against a broad set of baselines, covering  PRMs of different sizes and UQ methods. 


\myparagraph{Small PRMs (1.5B):} we use 2 PRMs fine-tuned from Qwen2.5-Math-1.5B: \texttt{Skywork-PRM-1.5B} \cite{skyworkopeno12024} and \texttt{H4-Qwen2.5-PRM-1.5B-0.2} \cite{huggingfaceh4_qwen2.5_math1.5b_instruct_prm0.2}.

\myparagraph{Large PRMs (7--8B):} we use (1) \texttt{Math-Shepherd-PRM-7B} \cite{wang-etal-2024-math}, which determines the process labels for each step by estimating the empirical probability of reaching the correct final answer (MC estimation); (2) \texttt{RLHFlow-PRM-8B-DeepSeek/Mistral} models \citep{xiong2024implementation} trained with MC-estimated labels from DeepSeek/Mistral rollouts; (3) \texttt{Universal-PRM-7B} \cite{auroraprm2025} trained using ensemble prompting and reverse verification; (4) \texttt{Qwen2.5-Math-7B-PRM800k} trained on the PRM800k dataset \cite{lightman2023letsverifystepstep}; and (5) \texttt{Qwen2.5-Math-PRM-7B} \citep{zhang2025lessonsdevelopingprocessreward}, which combines MC estimation with LLM-as-a-Judge consensus and currently achieves the best result on ProcessBench compared to PRMs of similar scale and computation \citep{zhao2025genprm}.

\myparagraph{UQ methods} evaluated in our experiments fall into two categories, differing in computational cost: (1) \emph{lightweight scores that use only single generation}: \texttt{Maximum Sequence Probability}, \texttt{Mean Token Entropy}, \texttt{Perplexity} \citep{fadeeva-etal-2023-lm}, \texttt{P(True)} \citep{kadavath2022languagemodelsmostlyknow}, \texttt{CCP} \citep{fadeeva2024fact}, and \texttt{Self-Certainty} \citep{kang2025scalable}, which show significant advantages in reasoning chain selection \citep{fu2025deepthinkconfidence}; and (2) \emph{sampling-based scores}: \texttt{Semantic Entropy}, \texttt{Lexical Similarity} \citep{kuhn2023semantic}, and \texttt{Degree Matrix} \citep{lin2024generating}. To compute the latter, we draw $M=10$ alternative steps per step position, making them much less computationally efficient. All methods are implemented using the LM-Polygraph framework \citep{fadeeva-etal-2023-lm,vashurin2024benchmarking}.

\subsection{Test Dataset Details} \label{appendix:test_set_details}

The mathematical domain includes \texttt{MATH} (high school and competition-style math problems; \citealp{hendrycks2021measuringmathematicalproblemsolving}), \texttt{GSM8K} (grade-school math word problems; \citealp{cobbe2021trainingverifierssolvemath}), and \texttt{ProofNet} (natural language proofs of undergraduate-level math problems; \citealp{azerbayev2023proofnetautoformalizingformallyproving}). 
Planning domain includes \texttt{NaturalPlan} \citep{zheng2024naturalplanbenchmarkingllms} (spans three real-world tasks -- trip planning, meeting planning, and calendar scheduling). The QA domain includes \texttt{ScienceQA} without multi-modal context \citep{lu2022learn} (covers 26 science subjects from elementary to high school) and \texttt{StrategyQA} \citep{geva2021strategyqa} -- a general knowledge QA benchmark that requires implicit multi-step reasoning.
Due to computational limitations, we evaluate on randomly sampled subsets of the test sets. 

\cref{tab:test_dataset_stats} presents the statistics of the test datasets used for the step-level benchmark. Importantly, we rely on DeepSeek-R1 for step-level evaluation, which is an expensive API-based LLM. We thus take a subset of test sets for this evaluation. While there are only a few hundred questions, there are a substantial number of reasoning steps, which allows us to draw representative insights.
\begin{table*}[h]

\centering
\footnotesize

\resizebox{0.75\linewidth}{!}{%
\begin{tabular}{l|l|cccc|ccc|c}
\toprule
\textbf{Dataset} & \textbf{Model} & \textbf{\#Questions} & \textbf{\#Steps} & \textbf{\multirowcell{Mean \\ \#Steps}} & \textbf{\multirowcell{\%Correct\\Steps}} & \textbf{\multirowcell{Answer Mean \\ Len (Tokens)}} \\
\midrule
\multirow{2}{*}{MATH} & Qwen3-8B & 200 & 1203 & 6.2 & 85.5\%  & 204 \\
 & Phi-4 & 200 & 2296 & 11.5 & 89.2\% & 417 \\
\midrule
\multirow{2}{*}{GSM8k} & Qwen3-8B & 200 & 1056 & 5.1 & 94.0\%  & 139 \\
 & Phi-4 & 200 & 1286 & 6.4 & 95.7\% & 174 \\
\midrule
\multirow{2}{*}{ProofNet} & Qwen3-8B & 186 & 1211 & 6.5 & 81.9\% & 236 \\
 & Phi-4 & 186 & 2227 & 12.0 & 91.1\% & 515 \\
\midrule
\multirow{2}{*}{Trips} & Qwen3-8B & 320 & 5492 & 17.2 & 46.9\%  &  641 \\
 & Phi-4 &  320 & 4486 & 14.0 & 44.87\%  & 584 \\
\midrule
\multirow{2}{*}{Meetings} & Qwen3-8B &  200 & 2569 & 12.8 & 40.9\%  &  739 \\
 & Phi-4 & 200 & 3350 & 16.75 & 52.3\%  & 584\\
\midrule
\multirow{2}{*}{Calendar} & Qwen3-8B &  200 & 1616 & 8.1 & 52.5\%  &  365  \\
 & Phi-4 &   200 & 2172 & 10.86 & 66.9\%  & 428\\
\midrule
\multirow{2}{*}{StrategyQA} & Qwen3-8B & 500 & 2865 & 5.73 & 81.1\% & 121 \\
 & Phi-4 & 500 & 3288 & 6.6 & 82.6\% & 182 \\
\midrule
\multirow{2}{*}{ScienceQA} & Qwen3-8B & 500 & 2621 & 5.24 & 88.7\% & 110 \\
 & Phi-4 & 500 & 3415 & 6.8 & 91.9\% & 172 \\
\bottomrule
\end{tabular}
}
\caption{Test dataset statistics used for step-level benchmark.}
\label{tab:test_dataset_stats}
\end{table*}

\subsection{Test Dataset Examples}
\label{appendix:dataset_examples}
\cref{tab:qa-examples} presents example datapoints randomly picked from our benchmarking datasets.

\newcolumntype{S}{>{\hsize=.65\hsize\raggedright\arraybackslash}X} 
\newcolumntype{B}{>{\hsize=1.35\hsize\raggedright\arraybackslash}X}

\begin{table*}[h]

\centering
\small
\begin{tabularx}{\textwidth}{@{}l | S | B@{}}
\toprule
\textbf{Dataset} & \textbf{Question} & \textbf{Answer} \\
\midrule

GSM8k &
\begin{minipage}[t]{.65\linewidth}\ttfamily\tiny
Natalia sold clips to 48 of her friends in April, and then she sold half as many clips in May. How many clips did Natalia sell altogether in April and May?
\end{minipage}
&
\begin{minipage}[t]{1.35\linewidth}\ttfamily\tiny
- Step 1: Natalia sold 48 clips in April.\\
- Step 2: Natalia sold half as many clips in May, which is $48 \div 2 = 24$.\\
- Step 3: Add them: $48 + 24 = 72$.\\
\textless Answer\textgreater: 72
\end{minipage}
\\
\midrule

ProofNet &
\begin{minipage}[t]{.65\linewidth}\ttfamily\tiny
If $r$ is rational ($r \neq 0$) and $x$ is irrational, prove that $r + x$ is irrational.
\end{minipage}
&
\begin{minipage}[t]{1.35\linewidth}\ttfamily\tiny
- Step 1: Assume for contradiction that $r + x$ is rational.\\
- Step 2: Then $x = (r + x) - r$ is rational.\\
- Step 3: This contradicts $x$ being irrational.\\
- Step 4: Therefore $r + x$ is irrational.\\
\textless Answer\textgreater: $r + x$ is irrational
\end{minipage}
\\
\midrule

MATH &
\begin{minipage}[t]{.65\linewidth}\ttfamily\tiny
Between what two consecutive integers is $\sqrt{30} + \sqrt{50}$ located? Answer in the form ``m and n''.
\end{minipage}
&
\begin{minipage}[t]{1.35\linewidth}\ttfamily\tiny
- Step 1: $\sqrt{30} \approx 5.477$.\\
- Step 2: $\sqrt{50} \approx 7.071$.\\
- Step 3: Sum $\approx 12.548$.\\
- Step 4: Lies between 12 and 13.\\
\textless Answer\textgreater: 12 and 13
\end{minipage}
\\
\midrule

Trips &
\begin{minipage}[t]{.65\linewidth}\ttfamily\tiny
You plan to visit 3 European cities for 14 days.\\
Stay in Istanbul for 5 days, Tallinn for 4 days, Zurich for 7 days.\\
From day 1-7, there is a show in Zurich.\\
Direct flights exist: Istanbul--Tallinn, Zurich--Tallinn, Zurich--Istanbul.\\
Find a valid trip plan.
\end{minipage}
&
\begin{minipage}[t]{1.35\linewidth}\ttfamily\tiny
- Step 1: Start in Zurich for days 1-7 (show).\\
- Step 2: Fly Zurich $\to$ Tallinn, stay days 8-11.\\
- Step 3: Fly Tallinn $\to$ Istanbul, stay days 12-14.\\
\textless Answer\textgreater: Day 1-7: Zurich, Day 8-11: Tallinn, Day 12-14: Istanbul
\end{minipage}
\\
\midrule

Meetings &
\begin{minipage}[t]{.65\linewidth}\ttfamily\tiny
You arrive at Financial District at 9:00AM.\\
Richard: Marina District 3:30-5:30PM (need 90 min).\\
Andrew: Alamo Square 5:00-10:00PM (need 120 min).\\
Travel: FD-Marina 15min, FD-Alamo 17min, Marina-Alamo 15min.\\
Plan a schedule to meet both.
\end{minipage}
&
\begin{minipage}[t]{1.35\linewidth}\ttfamily\tiny
- Step 1: Leave FD 2:45PM, arrive Marina 3:00PM.\\
- Step 2: Meet Richard 3:30-5:00PM.\\
- Step 3: Leave Marina 5:15PM, arrive Alamo 5:30PM.\\
- Step 4: Meet Andrew 5:30-7:30PM.\\
\textless Answer\textgreater: FD $\to$ Marina $\to$ Alamo schedule works
\end{minipage}
\\
\midrule

Calendar &
\begin{minipage}[t]{.65\linewidth}\ttfamily\tiny
Schedule a 30min meeting for Laura and Paul on Monday between 9:00-17:00.\\
Laura busy: 11:30-12:30, 14:30-15:00, 16:00-16:30.\\
Paul busy: 9:30-10:00, 11:00-14:30, 15:00-17:00.\\
Paul prefers not after 9:30.
\end{minipage}
&
\begin{minipage}[t]{1.35\linewidth}\ttfamily\tiny
- Step 1: Laura free 9:00-11:30.\\
- Step 2: Paul free 9:00-9:30 and before 9:30 only.\\
- Step 3: Intersection is 9:00-9:30.\\
\textless Answer\textgreater: Meeting 9:00-9:30
\end{minipage}
\\
\midrule

StrategyQA &
\begin{minipage}[t]{.65\linewidth}\ttfamily\tiny
Did Donatello use a smartphone?
\end{minipage}
&
\begin{minipage}[t]{1.35\linewidth}\ttfamily\tiny
- Step 1: Donatello lived 1386-1466.\\
- Step 2: Smartphones invented 21st century (iPhone 2007).\\
- Step 3: Impossible for him to have used one.\\
\textless Answer\textgreater: No, he did not
\end{minipage}
\\
\midrule

ScienceQA &
\begin{minipage}[t]{.65\linewidth}\ttfamily\tiny
Which word is not like the others?\\
A. horse \\ B. goat \\ C. squirrel \\ D. leg
\end{minipage}
&
\begin{minipage}[t]{1.35\linewidth}\ttfamily\tiny
- Step 1: Horse, goat, squirrel are animals.\\
- Step 2: Leg is a body part.\\
- Step 3: Outlier is D.\\
\textless Answer\textgreater: D. leg
\end{minipage}
\\
\bottomrule
\end{tabularx}
\caption{Examples of question-answer pairs from each dataset, ordered by domain: Math (GSM8k, ProofNet, MATH), Planning (Trips, Meetings, Calendar), and QA (StrategyQA, ScienceQA).}
\label{tab:qa-examples}
\end{table*}


\section{Human Validation of LLM-as-a-Judge} \label{appendix:human_validation}



DeepSeek-R1's judgement for step-level correctness is compared against human annotations. The results are shown in \cref{tab:steps_acc}. Besides the 17K PRM800K labels, we manually annotate 1000 steps spanning all studied domains. Here are the sampling details.

We first randomly shuffle the DeepSeek-R1 annotated test set. The human annotation then starts from the first row of the shuffled dataset. Human annotators are not provided with DeepSeek-R1's rationales (i.e., contents within the \texttt{<think>} tag) and must independently assess the correctness of each reasoning step, following the step-level correctness definition in \cref{fig:anno_prompt_step_level}.

Regardless of dataset difficulty, annotators are requested to annotate at least 100 steps from each dataset, with the option to annotate more depending on their availability.

\begin{table}[h]
\centering
\scriptsize
\begin{tabular}{lcc}
\toprule
\textbf{Dataset} & \textbf{\# Steps} & \textbf{Acc.} \\
\midrule
PRM800K  & 17,067 & 95.29 \\
ProofNet & 193 & 87.05 \\
Trips    & 102 & 93.06 \\
Meetings & 101 & 95.70 \\
Calendar & 102 & 91.09 \\
StrQA    & 313 & 95.85 \\
SciQA    & 245 & 99.28 \\
\bottomrule
\end{tabular}
\caption{
Accuracy of the DeepSeek-R1-based step-level evaluation pipeline relative to human labels.}
\label{tab:steps_acc}
\end{table}





\section{Theoretical Analysis of ReProbe Efficiency} \label{appendix:reprobe_efficiency}

Let the target LLM generate a reasoning trace of total length $T$ tokens, segmented into $S$ reasoning steps with token lengths $\{m_s\}_{s=1}^S$ such that $\sum_{s=1}^S m_s = T$, and define $m_{\max} \coloneqq \max_s m_s$.
A PRM is a separate Transformer critic with $L_{\text{prm}}$ layers, hidden width $d_{\text{prm}}$, and parameter count $P_{\text{prm}}$, which scores (possibly growing) text prefixes.
With standard dense self-attention, evaluating the PRM over a length-$T$ sequence incurs time at least $\Omega\!\big(L_{\text{prm}}\,d_{\text{prm}}\,T^2\big)$ (quadratic attention) and memory $\mathcal{O}\!\big(P_{\text{prm}} + L_{\text{prm}}\,T\,d_{\text{prm}}\big)$ (parameters plus KV-cache).

Hidden-states ReProbe instead taps per-token hidden representations from the frozen target LLM.
Let $f$ be the probe’s per-token feature dimension.
ReProbe is a lightweight Transformer with $L_{\text{rp}}$ layers, hidden width $d_{\text{rp}}$, and parameter count $P_{\text{rp}}$, applied \emph{within each step} to the length-$m_s$ feature sequence, preceded by a token-wise linear projection $\mathbb{R}^f \!\to\! \mathbb{R}^{d_{\text{rp}}}$.
ReProbe's total time complexity ($TC_{rp}$) over the full trace:
\begin{equation}
\begin{aligned}
TC_{rp}
&= \mathcal{O}\!\Big(
\sum_{s=1}^S \![m_s f d_{\text{rp}}
+ L_{\text{rp}}(m_s^2 d_{\text{rp}} + m_s d_{\text{rp}}^2)]
\Big) \\
&= \mathcal{O}\!\Big(
T f d_{\text{rp}}
+ L_{\text{rp}}(
d_{\text{rp}}\!\textstyle\sum_s m_s^2
+ d_{\text{rp}}^2 T)
\Big).
\end{aligned}
\end{equation}

where $m_s f d_{\text{rp}}$ is the cost of the per-token feature projection, $m_s^2 d_{\text{rp}}$ is the quadratic self-attention mixing cost within a step, and $m_s d_{\text{rp}}^2$ accounts for the token-wise dense linear maps inside each Transformer block (e.g., Q/K/V and output projections, as well as the MLP).
Its memory is $\mathcal{O}\!\big(P_{\text{rp}} + m_{\max}(f + L_{\text{rp}} d_{\text{rp}})\big)$ when step-local features/KV-cache are discarded after scoring each step.
Using $\sum_s m_s^2 \le m_{\max} T$, we obtain the upper bound
\[
TC_{rp}
\le
\mathcal{O}\!\left(
T f d_{\text{rp}} + L_{\text{rp}}(d_{\text{rp}} m_{\max} T + d_{\text{rp}}^2 T)
\right),
\]
which is near-linear in $T$ whenever $m_{\max} \ll T$ (equivalently, $S$ grows with $T$).
By contrast, PRM-based verification remains at least quadratic in $T$ due to attention over length-$T$ prefixes and additionally incurs substantially larger memory complexity (typically $P_{\text{rp}}\!\ll\!P_{\text{prm}}$).

\section{Prompts} \label{appendix:prompt}

This section collects the prompts used in our experiments for reasoning generation and for step- and chain-level annotation.

\subsection{Prompts for Reasoning}
\label{appendix:llm_prompt}

For LLMs in non-thinking mode, we use a domain-agnostic, format-enforcing prompt to elicit structured step-by-step reasoning, detailed in \cref{fig:llm_prompt_template}. For thinking mode, we use a slightly different prompt specified in \cref{fig:free_form_llm_prompt}.

\begin{figure*}[h]
\centering
\begin{minipage}{\linewidth}
\begin{lstlisting}
<|im_start|>user
You will be presented with a <Question>. Before providing the [Answer], you should first think step-by-step carefully.

Your response format:
<start of response>
Reasoning Steps:
- Step 1: [Your first reasoning step]
- Step 2: [Your second reasoning step]
- Step 3: [Next step, and so on...]
...
- Step N: [Final reasoning step]
<Answer>: [Your final answer]
<end of response>

Strict Requirements:
- DO NOT include any text outside the specified format.
- Each reasoning step MUST be written on a **single line only**: NO line breaks, bullet points, or substeps within a step.
- Each step should express one precise and **self-contained** logical operation, deduction, calculation, or fact application.
- Steps MUST provide explicit result of the step or concrete reasoning outcomes. Avoid vague explanations or meta-descriptions of the reasoning process.
    - For example:
        - Good: "- Step 1: Multiply 5 by 4, which equals 20."
        - Bad: "- Step 1: Multiply 5 by 4." (no result of the step or concrete reasoning outcome)
- Continue writing steps until the problem is solved.
- Violating ANY requirement above is NOT acceptable.

Now answer:
<Question>: {q}<|im_end|>
<|im_start|>assistant
<think>

</think>
Reasoning Steps:
\end{lstlisting}
\end{minipage}
\caption{Prompt template for Phi-4 and Qwen3-8B in non-thinking mode. The prompt is designed to elicit structured step-by-step reasoning.}
\label{fig:llm_prompt_template}
\end{figure*}

\begin{figure*}[h]
\centering
\begin{minipage}{\linewidth}
\begin{lstlisting}
<|im_start|>user
Answer the following question. Enclose your answer in <answer></answer>.
<Question>: {q}<|im_end|>
<|im_start|>assistant

\end{lstlisting}
\end{minipage}
\caption{Prompt template for Qwen3-1.7B and Qwen3-32B in native thinking mode.}
\label{fig:free_form_llm_prompt}
\end{figure*}

\subsection{Training Data Annotation Prompts}
\label{appendix:anno_prompt}

For step-level annotation, we use the 2-stage prompts shown in \cref{fig:anno_prompt_step_level}, while chain-level correctness annotations are obtained using the prompt in \cref{fig:anno_prompt_chain_level}. Since the native thinking mode generates much longer CoT than in the non-thinking mode, we use a different annotation prompt. Specifically, rather than annotating all steps in one LLM call, we annotate one step per LLM call, using the question, ground truth, and previously generated steps as input. \cref{fig:free_form_anno_prompt} shows the annotation prompt template.

\begin{figure*}[h]
\centering
\begin{minipage}{\linewidth}
\small
\textbf{Step-Level Annotation – Stage 1 Prompt:}
\begin{lstlisting}
You are given a problem, a ground-truth solution, and a step-by-step student solution. Your task is to analyze each step in the student's solution to determine whether it is both logically correct and relevant.

Instructions:
- Carefully examine each student step for logical errors or unnecessary/redundant reasoning.
- If all steps are correct and they lead to the same final answer as the ground-truth solution, conclude that there are no errors.
- If any step contains an error that would prevent the student from reaching the correct solution, identify and report those specific steps with an explanation.

PROBLEM:
{problem}

GROUND-TRUTH SOLUTION:
{answer}

STUDENT'S SOLUTION STEPS:
{steps}

Now, please evaluate whether the student's steps are correct and logical.
\end{lstlisting}

\vspace{0.8em}
\textbf{Step-Level Annotation – Stage 2 Prompt (Postprocessing):}
\begin{lstlisting}
You are given:
- A problem
- A student's step-by-step solution (as a Python list of string steps)
- An assessment of student's solution

Your task:
Output a single Python list where each element is:
- 1 if the corresponding step is correct
- 0 if the step is incorrect

Important:
- Output only the list, nothing else.
- The list must have the same length as the number of steps.

PROBLEM:
{problem}

STUDENT'S SOLUTION STEPS:
{steps}

ASSESSMENT OF STUDENT SOLUTION STEPS:
{reply}

OUTPUT LIST:
\end{lstlisting}
\end{minipage}
\caption{Two-step prompting procedure for step-level correctness annotation. The first stage evaluates the solution and identifies flaws, while the second converts this into binary correctness labels.}
\label{fig:anno_prompt_step_level}
\end{figure*}

\begin{figure*}[h]
\centering
\begin{minipage}{\linewidth}
\small
\begin{lstlisting}
You will be given a <Problem> and its proposed <Solution>. Your task is to assess whether the solution is **correct** or **incorrect**.

Respond using the **exact format** below, do not include any text outside this template.
Output format:
<start of response>
Solution comments:
... your comments on the solution, explaining reasoning, pointing out any errors or confirming correctness ...
<Grade>: (Correct|Incorrect)
<end of response>

<Problem>: {problem}

<Solution>: {solution}
\end{lstlisting}
\end{minipage}
\caption{The prompt used for annotating chain-level correctness by evaluating the full reasoning trace.}
\label{fig:anno_prompt_chain_level}
\end{figure*}

\begin{figure*}[h]
\centering
\begin{minipage}{\linewidth}
\begin{lstlisting}
You are given a problem, its ground-truth solution, and a student's (incomplete) reasoning process so far.
Your task is to determine whether the **latest step** in the student's reasoning is correct.

Instructions:
- A step is **wrong** if it contains explicit logical or computational errors, or if it contradicts any previous steps.
- Redundant, unnecessary, or non-informative steps are **not** considered wrong.
- If the latest step is correct, output **1**. If it is wrong, output **0**.
- Respond with the number (0 or 1) **only**, with no extra text or explanation.

PROBLEM:
{problem}

GROUND-TRUTH SOLUTION:
{answer}

STUDENT'S REASONING PROCESS SO FAR:
{steps}

THE LATEST STEP TO EVALUATE:
{latest_step}
\end{lstlisting}
\end{minipage}
\caption{Prompt template for annotating correctness of reasoning steps in native thinking mode.}
\label{fig:free_form_anno_prompt}
\end{figure*}

\section{The Usage of LLMs} 
In this paper, we mainly use LLMs as objects of study. We also use DeepSeek-R1 and Qwen3-8B as training data annotators, as detailed in \cref{sec:method} and \cref{appendix:llm_prompt}. We also use DeepSeek-R1 as a judge to grade answer correctness, as detailed in \cref{sec:exp_setup}. DeepSeek-R1's accuracy in these tasks is manually validated through human annotation (see \cref{tab:steps_acc}). In coding and writing, we use LLM assistants (e.g., ChatGPT) to identify grammar errors and debug. Such usage is under careful human supervision. 

\end{document}